# *Mining the Web for Lexical Knowledge to Improve Keyphrase Extraction: Learning from Labeled and Unlabeled Data*

P.D. Turney

August 13, 2002

# *Mining the Web for Lexical Knowledge to Improve Keyphrase Extraction: Learning from Labeled and Unlabeled Data*

P.D. Turney

August 13, 2002

## *Contents*

*Abstract*

A journal article is often accompanied by a list of *keyphrases*, composed of about five to fifteen important words and phrases that capture the article's main topics. Keyphrases are useful for a variety of purposes, including summarizing, indexing, labeling, categorizing, clustering, highlighting, browsing, and searching. The task of automatic keyphrase extraction is to select keyphrases from within the text of a given document. Automatic keyphrase extraction makes it feasible to generate keyphrases for the huge number of documents that do not have manually assigned keyphrases. Good performance on this task has been obtained by approaching it as a supervised learning problem. An input document is treated as a set of candidate phrases that must be classified as either keyphrases or non-keyphrases. To classify a candidate phrase as a keyphrase, the most important features (attributes) appear to be the frequency and location of the candidate phrase in the document. Recent work has demonstrated that it is also useful to know the frequency of the candidate phrase as a manually assigned keyphrase for other documents in the same domain as the given document (e.g., the domain of computer science). Unfortunately, this *keyphrase-frequency* feature is *domain-specific* (the learning process must be repeated for each new domain) and *training-intensive* (good performance requires a relatively large number of training documents in the given domain, with manually assigned keyphrases). The aim of the work described here is to remove these limitations. In this paper, I introduce new features that are conceptually related to *keyphrase-frequency* and I present experiments that show that the new features result in improved keyphrase extraction, although they are neither domain-specific nor training-intensive. The new features are generated by issuing queries to a Web search engine, based on the candidate phrases in the input document. The feature values are calculated from the number of hits for the queries (the number of matching Web pages). In essence, these new features are derived by mining lexical knowledge from a very large collection of unlabeled data, consisting of approximately 350 million Web pages without manually assigned keyphrases.

## 1. Introduction

A journal article is often accompanied by a list of *keyphrases*, composed of about five to fifteen important words and phrases that express the primary topics and themes of the paper. For an individual document, keyphrases can serve as a highly condensed summary, they can supplement or replace the title as a label for the document, or they can be highlighted within the body of the text, to facilitate speed reading (skimming). For a collection of documents, keyphrases can be used for indexing, categorizing (classifying), clustering, browsing, or searching. Keyphrases are most familiar in the context of journal articles, but many other types of documents could benefit from the use of keyphrases, including Web pages, email messages, news reports, magazine articles, and business papers.

The vast majority of documents currently do not have keyphrases. Although the potential benefit is large, it would not be practical to manually assign keyphrases to them. This is the motivation for developing algorithms that can automatically supply keyphrases for a document. There are two general approaches to this task: *keyphrase assignment* and *keyphrase extraction*. Both approaches use supervised machine learning from examples. In both cases, the training examples are documents with manually supplied keyphrases. Otherwise, the two approaches are quite different.

In keyphrase assignment, there is a predefined list of keyphrases (in the terminology of library science, a *controlled vocabulary* or *controlled index terms*). These keyphrases are treated as classes, and techniques from text classification (text categorization) are used to learn models for assigning a class to a given document (Leung and Kan, 1997; Dumais *et al.*, 1998). A document is converted to a vector of features and machine learning techniques are used to induce a mapping from the feature space to the list of keyphrases. The features are based on the presence or absence of various words or phrases in the input documents. Usually a document may belong to several different classes. That is, a learned model will map an input document to several different controlled vocabulary keyphrases.

In keyphrase extraction, keyphrases are selected from within the body of the input document, without a predefined list. When authors assign keyphrases without a controlled vocabulary (in library science, *free text keywords* or *free index terms*), typically about 70% to 80% of their keyphrases appear somewhere in the body of their documents (Turney, 1997, 1999, 2000). This suggests the possibility of using author-assigned free text keyphrases to train a keyphrase extraction system. In this approach, a document is treated as a set of candidate phrases and the task is to classify each candidate phrase as either a keyphrase or non-keyphrase (Turney, 1997, 1999, 2000; Frank *et al.*, 1999; Witten *et al.*, 1999, 2000). A feature vector is calculated for each candidate phrase and machine learning techniques are used to learn a model that can classify a phrase as a keyphrase or non-keyphrase. The features include the frequency and location of the candidate phrase in the input document. The features can also be based on information that is external to the given input document.

A limitation of keyphrase assignment is that it must be retrained every time a new phrase is added to the controlled vocabulary. Keyphrase extraction does not require retraining, but it can only supply keyphrases that appear somewhere in the input document, unlike keyphrase assignment, which does not have this limitation.

A learning algorithm is *training-intensive* when it requires a relatively large amount of labeled training examples in order to perform well. Keyphrase assignment is training-intensive when the controlled vocabulary is large, since there must be several training example documents for each keyphrase in the vocabulary. On the other hand, keyphrase extraction typically works well with only about 50 training documents (Turney, 1997, 1999, 2000; Frank *et al.*, 1999; Witten *et al.*, 1999, 2000).

A learning algorithm is *domain-specific* when the learned model does not generalize well from one domain to another domain. Keyphrase assignment is domain-specific, since the appropriate controlled vocabulary will vary from one domain to another. For example, the vocabulary of physics articles is distinct from the vocabulary of computer science articles. On the other hand, keyphrase extraction performs well when trained on articles from one domain and then tested on articles from a completely different domain (Turney, 1997, 1999, 2000; Frank *et al.*, 1999; Witten *et al.*, 1999, 2000).

Section 2 discusses prior work on automatic keyphrase extraction. In the GenEx (Turney, 1997, 1999, 2000) and Kea (Frank *et al.*, 1999; Witten *et al.*, 1999, 2000) automatic keyphrase extraction systems, the most important features for classifying a candidate phrase are the frequency and location of the phrase in the document. There are several versions of Kea, using various methods for finding candidate phrases in a document and various features for classifying the candidate phrases. In one version of Kea, the frequency and location features are supplemented with a feature based on the frequency of the candidate phrase as a manually assigned keyphrase for other documents in the same domain as the given document (Frank *et al.,* 1999). This new feature is called *keyphrase-frequency*. The experimental results show a significant improvement in keyphrase extraction when the new feature is added (Frank *et al.,* 1999).

The *keyphrase-frequency* feature seems to be very useful for keyphrase extraction, but it has two important limitations: it is *domain-specific* and *training-intensive*. The training process must be repeated for each new domain and the training process requires a relatively large number of labeled

*1. Introduction*

training examples to perform well. Suppose that Kea has been trained for the domain of computer science, but now we have a document from the domain of physics. It will not help us to know that a phrase such as "distributed computing" has a high *keyphrase-frequency* in the domain of computer science. This particular phrase is not likely to be appropriate as a keyphrase for a physics paper. To achieve good performance with physics documents, *keyphrase-frequency* must be calculated using physics documents (see the experiment in Section 5).

In a sense, when Kea is supplemented with the *keyphrase-frequency* feature, it becomes a kind of hybrid of the two approaches, keyphrase extraction and keyphrase assignment. The list of phrases for which *keyphrase-frequency* is non-zero is somewhat like the controlled vocabulary that is used in keyphrase assignment. The problems of domain-specificity and training-intensiveness that come with the new *keyphrase-frequency* feature are classical problems with the keyphrase assignment approach.

Section 3 introduces new features that are inspired by *keyphrase-frequency*, yet are neither domain-specific nor training-intensive. These new features exploit the Web as a source of unlabeled data (documents without manually assigned keyphrases). The new features are based on an unsupervised learning algorithm called PMI-IR (Turney, 2001, 2002). This algorithm uses Pointwise Mutual Information (PMI) to measure the strength of association between pairs of words. PMI is a statistical measure of word association, based on the frequency of co-occurrence of pairs of words. PMI-IR uses Information Retrieval (IR) to acquire the frequency information that is needed to calculate PMI. The new features are calculated by issuing queries to a Web search engine and analyzing the resulting number of hits (the number of matching Web pages). The queries use the candidate phrases from the input document. In essence, these new features are derived by mining lexical knowledge from a very large collection of unlabeled data, consisting of approximately 350 million Web pages without manually assigned keyphrases.

Beginning with Section 4, the following seven sections present a series of seven experiments. The experiments use the Kea system as a framework for comparing three sets of features: (1) the *baseline* feature set is calculated from the frequency and location of the candidate phrases in the input document, (2) the *keyphrase* feature set is the baseline feature set supplemented with the *keyphrase-frequency* feature, and (3) the *query* feature set is the baseline feature set supplemented with features that are calculated using queries to a Web search engine. The first experiment compares the three sets of features on the CSTR corpus (computer science papers from the Computer Science Technical Reports collection of the New Zealand Digital Library Project), when part of this corpus is used for training and the rest is used for testing. In this experiment, the keyphrase features perform best, followed by the query features, and lastly the baseline features. This experiment demonstrates that the query features can improve keyphrase extraction.

In the second experiment (Section 5), part of the CSTR corpus is used for training, but the LANL corpus (physics papers from the arXiv repository at the Los Alamos National Laboratory) is used for testing. In this case, the query features perform best, followed by the baseline features, and finally the keyphrase features. When the training domain does not correspond to the testing domain, the *keyphrase-frequency* feature becomes detrimental, instead of beneficial. However, the query features generalize well from the computer science domain to the physics domain, which shows that they are not domain-specific.

There are a total of twelve features in the query feature set. The third experiment (Section 6) looks at various subsets of the twelve features. The results suggest that all twelve features are useful.

Since the query features were inspired by *keyphrase-frequency* and they are conceptually similar, it was conjectured that the output of Kea with the query feature set would be similar to the output with the keyphrase feature set. The fourth experiment (Section 7) tests this conjecture and finds it to be false. The query and baseline features are relatively similar in behaviour, but the keyphrase features are substantially different from the other two.

This difference suggests that it could be beneficial to combine the query features with the keyphrase features, since they seem to be somewhat independent. The fifth experiment (Section 8) evaluates a hybrid of the two feature sets. When trained on part of the CSTR corpus and tested on another part of the CSTR corpus, the hybrid feature set performs better than the other feature sets. When trained on the CSTR corpus and tested on the LANL corpus, the performance of the hybrid feature set is very similar to the performance of the keyphrase feature set alone. Thus there is some benefit to combining the features, but the combined feature set has the same limitations as the keyphrase feature set: domain-specificity and training-intensiveness.

The first two experiments measured performance by the number of correct classifications; that is, by the level of agreement between Kea and the author. However, a candidate phrase might be a reasonable, appropriate keyphrase, even though it was not chosen by the author as a keyphrase. One test of reasonableness is whether the phrase was chosen by any author of any paper in the same domain. The sixth experiment (Section 9) repeats the setup of the first two experiments, but measures the performance of the different feature sets by the level of agreement between Kea and the manually assigned keyphrases of any paper in the same testing set. This experiment does not evaluate whether the phrases output by Kea are appropriate for the given input document; it evaluates whether the phrases seem reasonable for a paper in the given domain, without regard to the actual content of the paper, beyond its domain. When trained and tested on the CSTR corpus, it is not surprising that the keyphrase feature set performs very well by this measure. The difference between the baseline features and the query features is small. When trained on the CSTR corpus and tested on the LANL corpus, the query features perform best, followed by the keyphrase features, and lastly the baseline features.

The seventh and final experiment (Section 10) measures the performance of the keyphrases as query terms for searching. The reasoning is that good keyphrases should be specific enough that they can be used to find the source document, from which they were taken, yet they should be general enough that they can also find many other, related documents from the same domain. This experiment compares the keyphrases extracted with the three different feature sets and also the authors' keyphrases. On the CSTR corpus, there are no significant differences among the baseline, query, and author keyphrases, but the *keyphrase-frequency* keyphrases are significantly more general than the others. On the LANL corpus, the *keyphrase-frequency* keyphrases are again the most general. This shows that there is a systematic bias towards generality in the phrases that are chosen when using the *keyphrase-frequency* feature.

Section 11 discusses limitations and future work. The main limitation of the new query features is the time that is required to calculate them. Almost all of this time is taken up with network traffic between the machine that runs the learning algorithm and the machine that hosts the Web search engine. However, within ten years, the average desktop personal computer will have enough memory to store 350 million Web pages locally and enough processing power to search them very rapidly.

The main findings of this paper are summarized in Section 12. The new query features improve keyphrase extraction but they are neither domain-specific nor training-intensive. They do require a large amount of unlabeled data, which currently makes them slow to calculate, but improving hardware will solve this problem. Although the query features are conceptually related to the *keyphrase-frequency* feature, their behaviour is substantially different. In some applications, when the required training data are available, it may be beneficial to combine the query features with the keyphrase features.

## 2. *Related Work*

Automatic keyphrase extraction is related to many other technologies, such as automatic summarization (Luhn, 1958; Edmundson, 1969; Kupiec *et al.*, 1995), information extraction (Soderland and Lehnert, 1994), and keyphrase assignment and automatic indexing (Sparck Jones, 1973; Field, 1975; Leung and Kan, 1997; Dumais *et al.*, 1998). For a general overview of the related literature and the position of keyphrase extraction within this literature, see Turney (2000). The focus here is specifically on prior work in learning to extract keyphrases from text.

### 2.1 GenEx: Genitor and Extractor

GenEx has two components, the Genitor genetic algorithm (Whitley, 1989) and the Extractor[1] keyphrase extraction algorithm (Turney, 1997, 1999, 2000). Extractor takes a document as input and extracts a list of words and phrases as output. The output of Extractor is controlled by a dozen numerical parameters. The setting of these parameters is determined by a training process, during which Genitor searches through the parameter space for values that yield a high overlap between the keyphrases assigned by the authors and the phrases that are output by Extractor. After training, the best parameter values can be hardcoded in Extractor, and Genitor is no longer needed.

To measure the overlap between the machine's phrases and the author's phrases, it is necessary to decide when two phrases match. If an author assigns the phrase "Distributed Computation" to a document and the machine extracts the phrase "distributed computing", this should count as a match. GenEx and Kea both use the same approach for counting matches: phrases are normalized by converting them to lower case and stemming them (removing suffixes). Both use the Iterated Lovins stemming algorithm, which applies the Lovins stemming algorithm repeatedly, until the word stops changing (Lovins, 1968; Turney, 1997).

Extractor generates candidate phrases by looking through the input document for any sequence of one, two, or three consecutive words. The consecutive words must not be separated by punctuation and must not include any stop words (words such as "the", "of", "to", "and", "if", "he", etc.). Candidate phrases are normalized by converting them to lower case and stemming them.

GenEx has been described in detail elsewhere (Turney, 1999, 2000). In the context of this paper, most of the details of the GenEx algorithm are not important, but it is relevant to know what features are used to select a candidate phrase for output. GenEx does not explicitly represent candidate phrases with feature vectors, but at an abstract level, it may be described as using a set of ten features (Table 1).

GenEx allows the user to specify the desired number of output phrases. When the user requests *N* phrases, the ten features are used to calculate a score for every candidate phrase and the top *N* highest scoring phrases are output. In the current version of GenEx, *N* can range from 3 to 30. After a candidate phrase has been selected for output, the final step is to restore the suffix and the original pattern of capitalization.

Experiments show that GenEx performs well when it is trained on one domain and then tested on quite different domains (Turney, 1997, 1999, 2000). Good results are possible with as few as 50 training documents. The level of agreement between the phrases output by GenEx and the phrases assigned by the authors depends on the number of output phrases requested by the user. If the user asks for seven phrases, typically about 20% of the phrases output by GenEx will match the author's phrases. This is similar to the level of agreement among different humans, assigning keyphrases to the same document (Furnas *et al.*, 1987). This figure underestimates the quality of the phrases, since the other 80% of the phrases are often subjectively good, although they do not correspond with the author's choices. A more accurate picture is obtained by asking human readers to rate the quality of the machine's output. In a sample of 205 human readers rating keyphrases for 267 Web pages, 62% of the 1,869 phrases output by GenEx were rated as "good", 18% were rated as "bad", and 20% were left unrated (Turney, 2000). This suggests that about 80% of the phrases are acceptable (not "bad") to human readers, which is sufficient for many applications.

---

[1] Extractor is an Official Mark of the National Research Council of Canada. Patent applications have been submitted for Extractor.

Table 1: Features implicit in GenEx.

|   | Name of feature | Description of feature | Type of feature |
|---|---|---|---|
| 1 | *num_words_phrase* | The number of words in the candidate phrase | numerical |
| 2 | *first_occur_phrase* | The location in the document where the phrase first occurs | numerical |
| 3 | *first_occur_word* | The location in the document of the earliest occurring word in the phrase | numerical |
| 4 | *freq_phrase* | The frequency of the phrase in the document | numerical |
| 5 | *freq_word* | The frequency of the most frequent word in the phrase | numerical |
| 6 | *relative_length* | The length of the candidate phrase, relative to other candidates in the document | numerical |
| 7 | *document_length* | The number of words in the document | numerical |
| 8 | *proper_noun* | Is the phrase a proper noun, based on the capitalization pattern? | boolean |
| 9 | *final_adjective* | Is the last word in the phrase an adjective, based on the suffix? | boolean |
| 10 | *common_verb* | Does the phrase contain a common verb, based on a list of common verbs? | boolean |

Extractor (GenEx without Genitor) has been licensed to 16 companies as a module for embedding in products and services. The current version, Extractor 7.2, handles plain text, Web pages, and email messages in English, French, German, Spanish, Japanese, and Korean. It is written in C and has an Application Program Interface (API) to facilitate embedding in other software. Wrappers are available for Perl, Java, Visual Basic, and Python.

## 2.2 Kea: Baseline Feature Set

Kea generates candidate phrases in much the same manner as Extractor (Frank *et al.*, 1999; Witten *et al.*, 1999, 2000). Kea then uses the Naïve Bayes algorithm to learn to classify the candidate phrases (Domingos and Pazzani, 1997). In one version of Kea, candidate phrases are classified using only two features: *TF × IDF* and *distance* (Frank *et al.*, 1999; Witten *et al.*, 1999, 2000). In the following, I call this the *baseline* feature set.

*TF × IDF* (Term Frequency times Inverse Document Frequency) is commonly used in information retrieval to assign weights to terms in a document (van Rijsbergen, 1979). This numerical feature assigns a high value to a phrase that is relatively frequent in the input document (the *TF* component), yet relatively rare in other documents (the *IDF* component). In Kea, *TF × IDF* is calculated as follows (Frank *et al.*, 1999):

## 2. Related Work

$$TF(P, D) \times IDF(P, C) = \frac{\text{freq}_d(P, D)}{\text{size}_d(D)} \times -\log_2 \frac{\text{freq}_c(P, C)}{\text{size}_c(C)} \tag{1}$$

$TF(P, D)$ is the probability that phrase *P* appears in document *D*, estimated by counting the number of times that *P* occurs in *D*, $\text{freq}_d(P, D)$, and dividing by the number of words in *D*, $\text{size}_d(D)$. $IDF(P, C)$ is the negative log of the probability that phrase *P* appears in any document in corpus *C*, estimated by counting the number of documents in *C* that contain *P*, $\text{freq}_c(P, C)$, and dividing by the number of documents in *C*, $\text{size}_c(C)$.

    The *TF* component of $TF \times IDF$ in Kea corresponds to the *freq_phrase* feature in GenEx. In GenEx, I have found that *TF* without *IDF* works well for keyphrase extraction. It is likely that the *relative_length* feature in GenEx serves as a surrogate for *IDF*. Banko *et al.* (1999) discuss the use of $TF \times TL$ as an alternative to $TF \times IDF$, where *TL* is Term Length (the number of characters in the term). They argue that *TL* can replace *IDF* because longer words tend to have a lower frequency than shorter words. One advantage of *TL* over *IDF* is that *TL* is easier to calculate. Furthermore, *TL* is defined for phrases outside of the training corpus *C*, unlike *IDF*. In Kea, $\text{freq}_c(P, C)$ and $\text{size}_c(C)$ are incremented by one, to avoid taking the logarithm of zero, but *TL* does not require this kind of adjustment. Also, Kea must assign the same *IDF* value to all out-of-corpus phrases, but *TL* only assigns phrases the same value when they are the same length.

    The *distance* feature in Kea is, for a given phrase in a given document, the number of words that precede the first occurrence of the phrase, divided by the number of words in the document. This corresponds to *first_occur_phrase* in GenEx.

    In Kea, a candidate phrase with a capitalization pattern that indicates a proper noun is deleted; it is not considered for output. In GenEx, the *document_length* feature and the *proper_noun* feature tend to work together. During training, GenEx tends to learn to avoid proper nouns (phrases for which the *proper_noun* feature has the value *true*) for long documents, but allow them for short documents. I conjecture that Kea was mainly developed with a corpus of relatively long documents, for which it is best to suppress proper nouns.

    The $TF \times IDF$ and *distance* features are real-valued. Kea uses Fayyad and Irani's (1993) algorithm to discretize the features. This algorithm uses a Minimum Description Length (MDL) technique to partition the features into intervals, such that the entropy of the class is minimized with respect to the intervals and the information required to specify the intervals.

    The Naïve Bayes algorithm applies Bayes' formula to calculate the probability of membership in a class, using the ("naïve") assumption that the features are statistically independent. Suppose that a candidate phrase has the feature vector $\langle T, D \rangle$, where *T* is an interval of the discretized $TF \times IDF$ feature and *D* is an interval of the discretized distance feature. Using Bayes' formula and the independence assumption, we can calculate the probability that the candidate phrase is a keyphrase, $p(key|T, D)$, as follows (Frank *et al.*, 1999):

$$p(key|T, D) = \frac{p(T|key) \cdot p(D|key) \cdot p(key)}{p(T, D)} \tag{2}$$

In this equation, $p(T|key)$ is the probability that the discretized $TF \times IDF$ feature has a value in the interval *T*, given that the candidate phrase is actually a keyphrase. $p(D|key)$ is the probability that the *distance* feature has a value in the interval *D*, given that the candidate phrase is actually a keyphrase. $p(key)$ is the prior probability that the candidate phrase is a keyphrase (the probability when the values *T* and *D* are not known). $p(T, D)$ is a normalization factor, to make $p(key|T, D)$ range from 0 to 1. $p(T, D)$ is the probability of $\langle T, D \rangle$ when the class is not known. These probabilities can easily be estimated from the frequencies in the training data.

After training, $p(key|T, D)$ can be used to estimate the probability that a candidate phrase $\langle T, D \rangle$ is a keyphrase. Kea ranks each of the candidate phrases by the estimated probability $p(key|T, D)$ that they belong to the keyphrase class. If the user requests *N* phrases, then Kea gives the top *N* phrases with the highest estimated probability as output.

Experiments comparing GenEx to Kea with the baseline feature set have shown no significant difference in performance (Frank *et al.*, 1999). Other experiments have demonstrated that Kea can improve browsing in a digital library, by automatically generating a keyphrase index (Gutwin *et al.*, 1999), or by automatically generating hypertext links (Jones and Paynter, 1999). The Kea source code is available under the GNU General Public License.[2] The current version, Kea 2.0, is written in Java. In the following experiments, I used Kea 1.1.4, which is written in a combination of Java, Perl, and C.

### 2.3 Kea: Keyphrase Feature Set

In another version of Kea, candidate phrases are classified using three features: *TF × IDF*, *distance*, and *keyphrase-frequency* (Frank *et al.*, 1999). I call this the *keyphrase* feature set. For a phrase *P* in a document *D* with a training corpus *C*, the *keyphrase-frequency* is the number of times *P* occurs as an author-assigned keyphrase in all documents in *C* that are different from *D*. Like the other two features, *keyphrase-frequency* is discretized using Fayyad and Irani's (1993) algorithm. Suppose that a candidate phrase has the feature vector $\langle T, D, K \rangle$, where *T* is an interval of the discretized *TF × IDF* feature, *D* is an interval of the discretized *distance* feature, and *K* is an interval of the discretized *keyphrase-frequency* feature. Using Bayes' formula and assuming independence, we can calculate as follows (Frank *et al.*, 1999):

$$p(key|T, D, K) = \frac{p(T|key) \cdot p(D|key) \cdot p(K|key) \cdot p(key)}{p(T, D, K)} \tag{3}$$

$p(K|key)$ is the probability that the discretized *keyphrase-frequency* feature has a value in the interval *K*, given that the candidate phrase is actually a keyphrase. These probabilities are easily estimated from the training data.

The baseline and keyphrase feature sets have been empirically evaluated using the CSTR corpus (computer science papers from the Computer Science Technical Reports collection in New Zealand).[3] This corpus consists of PostScript files that have been collected from Web sites around the world. The documents are journal papers, conference papers, technical reports, and preprints in computer science. There is a fair amount of noise in the corpus, because the PostScript files were automatically converted to plain text. For each document in the corpus, the author's keyphrase list was removed from the title page and placed in a separate file.

In one experiment with the baseline features, the size of the training set was varied from 1 to 130 documents, while the size of the testing set was fixed at 500 documents (Section 2.4.2 in Frank *et al.*, 1999). The performance measure was the number of machine-extracted keyphrases that matched the author-assigned keyphrases. The performance on the testing set improved at first, but leveled off after about 50 training documents.

In another experiment, the *TF × IDF* and *distance* features were trained on 130 documents and a separate set of 100 to 1000 documents was used to train the *keyphrase-frequency* feature (Section 3.2 in Frank *et al.*, 1999). (Since the Naïve Bayes algorithm assumes that features are independent, there is no difficulty in training the features separately.) The baseline features (*TF × IDF* and distance) and the keyphrase features (*TF × IDF*, *distance*, and *keyphrase-frequency*) were then evaluated on the same testing set of 500 documents. The keyphrase feature set performed significantly better than the baseline feature set. The difference in performance rose steadily as the number of documents used to train *keyphrase-frequency* rose from 100 to 1000. It appears that the performance would have continued to rise with more than 1000 training documents for *keyphrase-frequency*.

---

2. See http://www.nzdl.org/Kea/.
3. See http://www.nzdl.org/.

*3. Removing Domain-Specificity: Mining for Lexical Knowledge*

As I mentioned in the introduction, the experiments show that the *keyphrase-frequency* feature improves keyphrase extraction, but this improvement comes with a cost: domain-specificity and training-intensiveness. The training process must be repeated for each new domain and requires a relatively large number of labeled training examples.

## 3. *Removing Domain-Specificity: Mining for Lexical Knowledge*

This section begins with an analysis of the *keyphrase-frequency* feature. In the first subsection, I argue that the assumptions underlying the *keyphrase-frequency* feature imply that keyphrases will tend to co-occur within a domain. This suggests that a statistical measure of co-occurrence might be used to replace the *keyphrase-frequency* feature. In the second subsection, I introduce a particular statistical measure of co-occurrence, PMI-IR (Turney, 2001). Finally, in the third subsection, I introduce a new set of features that are based on PMI-IR.

### 3.1 A Model of the Keyphrase-Frequency Feature

Consider the following simple model of how the *keyphrase-frequency* feature works. Suppose we have some documents with associated keyphrases, from the domains of computer science and physics. Let $C$ be the set of all keyphrases that are particularly good for computer science documents and let $P$ be the set of all keyphrases that are particularly good for physics documents. Let $D_C$ be a probability distribution on $C$ and let $D_P$ be a probability distribution on $P$. Imagine that we assign keyphrases to a document by deciding whether it is a computer science document or a physics document, and then selecting phrases randomly from the corresponding keyphrase set, $C$ or $P$, using the corresponding probability distribution, $D_C$ or $D_P$. If this is a reasonable first-order approximation of how authors assign keyphrases to their documents, then the *keyphrase-frequency* feature will work well. The *keyphrase-frequency* feature is simply an estimate of the probability distribution, $D_C$ or $D_P$, for a given domain, $C$ or $P$. In fact, if this simple model were completely sufficient to describe how authors assign keyphrases, then the keyphrase-frequency feature would be the only feature that worked well; the *TF × IDF* and *distance* features would be irrelevant.

A consequence of this model is that keyphrases will tend to co-occur within a domain. That is, computer science keyphrases will tend to occur together and physics keyphrases will tend to occur together. This is trivially true if $C$ and $P$ have an empty intersection, but it is even true when $C$ and $P$ are equal, as long as $D_C$ and $D_P$ are distinct. Suppose that we have a large collection of documents with associated keyphrases, generated according to our simple model. Let $C$ equal $P$ and let $D_C$ and $D_P$ be distinct probability distributions. Suppose that half of the collection consists of computer science papers and the other half is physics papers. Let us say that two keyphrases, $k_1$ and $k_2$, tend to co-occur if the probability that they are both assigned as keyphrases for the same document $p(k_1 \& k_2)$ is greater than the expected probability $p(k_1) \cdot p(k_2)$, assuming that they are independent. Let us say that a keyphrase $k$ is a *physics keyphrase* if it is more probable according to the distribution $D_P$ than according to $D_C$; otherwise, if it is more probable according to $D_C$, we will call it a *computer science keyphrase*. We can easily see that $k_1$ and $k_2$ will tend to co-occur if and only if they are both physics keyphrases or both computer science keyphrases. In other words, keyphrases will tend to co-occur within a domain.

The *keyphrase-frequency* feature has two limitations. It requires keyphrases to be provided for each document and it requires the domains of the documents to be explicitly identified (in our example, either physics or computer science). Let us consider an alternative model that does not have these limitations. Let $G$ be a set of general-purpose phrases (not only keyphrases) and let $D_G$ be a probability distribution on $G$. Suppose that a document in computer science is generated by randomly sampling phrases from $G$ according to $D_G$ and also phrases from $C$ according to $D_C$. Similarly, a document in physics is generated by randomly sampling phrases from $G$ according to $D_G$ and from $P$ according to $D_P$. Now let's take a document and run it through Kea, using the baseline feature set. Assume that we do not know whether it's a computer science document or a physics document. If the *IDF* component in the $TF \times IDF$ feature is based on a good estimate of $D_G$, and if the $D_C$ and $D_P$ distributions are significantly different from $D_G$, then the $TF \times IDF$ feature will tend to pick out phrases from $C$ or $P$, rather than $G$, because these phrases will tend to have a higher *TF* than we would expect from $D_G$. Thus the phrases that are output by Kea will tend to have a higher density of phrases from $C$ or $P$ than $G$, compared to the density in the input document.

This suggests the following two-pass algorithm. In the first pass, we use Kea with the baseline feature set. Then we take the $K$ top ranked phrases output by Kea. If Kea was successful, many of these $K$ phrases are from $C$ or $P$. By assumption, we do not know whether the input document is a computer science document or a physics document, but we do know that its keyphrases will tend to co-occur. Assume that we are more confident in the quality of the top $K$ phrases than the remaining phrases. In the second pass, we use the top $K$ phrases output by Kea to screen the remaining phrases, by measuring the degree of co-occurrence between the top $K$ phrases and the remaining phrases.

Thus we can exploit co-occurrence information without knowing the domain of the input document and without knowing the keyphrases that are assigned to the documents. Using this two-pass algorithm, we can overcome the two limitations of the keyphrase-frequency feature. The next subsection presents a measure of co-occurrence and the final subsection gives the details of the two-pass algorithm.

### 3.2  PMI-IR: Mining the Web for Synonyms

PMI-IR uses Pointwise Mutual Information (PMI) and Information Retrieval (IR) to measure the semantic similarity between pairs of words or phrases (Turney, 2001). The algorithm involves issuing queries to a search engine (the IR component) and applying statistical analysis to the results (the PMI component). The power of the algorithm comes from its ability to exploit a huge collection of text. In the following experiments, I used the AltaVista® search engine, which indexes about 350 million Web pages in English.[4]

PMI-IR was designed to recognize synonyms. The task of synonym recognition is, given a problem word and a set of alternative words, choose the member from the set of alternative words that is most similar in meaning to the problem word. PMI-IR has been evaluated using 80 synonym recognition questions from the Test of English as a Foreign Language (TOEFL) and 50 synonym recognition questions from a collection of tests for students of English as a Second Language (ESL). On both tests, PMI-IR scores 74% (Turney, 2001). For comparison, the average score on the 80 TOEFL questions, for a large sample of applicants to US colleges from non-English speaking countries, was 64.5% (Landauer and Dumais, 1997). Landauer and Dumais (1997) note that, "… we have been told that the average score is adequate for admission to many universities." Latent Semantic Analysis (LSA), another statistical technique, scores 64.4% on the 80 TOEFL questions (Landauer and Dumais, 1997).[5]

---

4. The AltaVista search engine is a service of the AltaVista Company of Palo Alto, California, http://www.altavista.com/. Including pages in languages other than English, AltaVista indexes more than 350 million Web pages, but these other languages are not relevant for answering questions in English. To estimate the number of English pages indexed by AltaVista, I used the Boolean query "the OR of OR an OR to" in the Advanced Search mode. The resulting number agrees with other published estimates. The primary reason for using AltaVista in the following experiments is the Advanced Search mode, which supports more expressive queries than many of the competing Web search engines.

PMI-IR is based on co-occurrence (Manning and Schütze, 1999). The core idea is that "a word is characterized by the company it keeps" (Firth, 1957). In essence, it is an algorithm for measuring the strength of associations among words (Turney, 2002).

Consider the following synonym test question, one of the 80 TOEFL questions. Given the problem word *levied* and the four alternative words *imposed, believed, requested, correlated,* which of the alternatives is most similar in meaning to the problem word? Let problem represent the *problem* word and $\{choice_1, choice_2, …, choice_n\}$ represent the alternatives. The PMI-IR algorithm assigns a score to each choice, $score(choice_i)$, and selects the choice that maximizes the score.

The PMI-IR algorithm is based on co-occurrence. There are many different measures of the degree to which two words co-occur (Manning and Schütze, 1999). PMI-IR uses Pointwise Mutual Information (PMI) (Church and Hanks, 1989; Church *et al.*, 1991), as follows:

$$\text{score}(choice_i) = \log_2\left(\frac{p(problem, choice_i)}{p(problem)p(choice_i)}\right) \quad (4)$$

Here, $p(problem, choice_i)$ is the probability that *problem* and $choice_i$ co-occur. If *problem* and $choice_i$ are statistically independent, then the probability that they co-occur is given by the product $p(problem)p(choice_i)$. If they are not independent, and they have a tendency to co-occur, then $p(problem, choice_i)$ will be greater than $p(problem)p(choice_i)$. Therefore the ratio between $p(problem, choice_i)$ and $p(problem)p(choice_i)$ is a measure of the degree of statistical dependence between *problem* and $choice_i$. The log of this ratio is the amount of information that we acquire about the presence of *problem* when we observe $choice_i$. Since the equation is symmetrical, it is also the amount of information that we acquire about the presence of $choice_i$ when we observe *problem*, which explains the term *mutual information*.[6]

Since we are looking for the maximum score, we can drop $\log_2$ (because it is monotonically increasing) and $p(problem)$ (because it has the same value for all choices, for a given problem word). Thus (4) simplifies to:

$$\text{score}(choice_i) = \frac{p(problem, choice_i)}{p(choice_i)} \quad (5)$$

In other words, each choice is simply scored by the conditional probability of the problem word, given the choice word, $p(problem|choice_i)$.

PMI-IR uses Information Retrieval (IR) to calculate the probabilities in (5). For the task of synonym recognition (Turney, 2001), I evaluated four different versions of PMI-IR, using four different kinds of queries. Only the first two versions of PMI-IR are needed here. The following description of these two different methods for calculating (5) uses the AltaVista® Advanced Search query syntax.[7] In the following, hits(*query*) represents the number of hits (the number of documents retrieved) given the query *query*.

---

5. This result for LSA is based on statistical analysis of about 30,000 encyclopedia articles. LSA has not yet been applied to text collections on the scale that can be handled by PMI-IR.
6. For an explanation of the term *pointwise* mutual information, see Manning and Schütze (1999).
7. See http://doc.altavista.com/adv_search/syntax.html.

1. In the simplest case, we say that two words co-occur when they appear in the same document:

$$\text{score}_1(choice_i) = \frac{\text{hits}(problem \text{ AND } choice_i)}{\text{hits}(choice_i)} \qquad (6)$$

   We ask the search engine how many documents contain both *problem* and $choice_i$, and then we ask how many documents contain $choice_i$ alone. The ratio of these two numbers is the score for $choice_i$.

2. Instead of asking how many documents contain both *problem* and $choice_i$, we can ask how many documents contain the two words close together:

$$\text{score}_2(choice_i) = \frac{\text{hits}(problem \text{ NEAR } choice_i)}{\text{hits}(choice_i)} \qquad (7)$$

   The AltaVista® NEAR operator constrains the search to documents that contain *problem* and $choice_i$ within ten words of one another, in either order.

When the queries yield a sufficient number of hits, $\text{score}_2$ tends to perform better than $\text{score}_1$, but the situation reverses when the queries return only a small number of hits, because $\text{hits}(problem \text{ NEAR } choice_i)$ is never larger than $\text{hits}(problem \text{ AND } choice_i)$.

Table 2 shows how $\text{score}_2$ is calculated for the sample TOEFL question, mentioned above. In this case, *imposed* has the highest score, so it is (correctly) chosen as the most similar of the alternatives for the problem word *levied*.

Table 2: Details of the calculation of $\text{score}_2$ for a sample TOEFL question.

| Query | | Hits |
|---|---|---|
| imposed | | 1,198,495 |
| believed | | 2,537,348 |
| requested | | 4,774,446 |
| correlated | | 244,353 |
| | | |
| levied NEAR imposed | | 3,593 |
| levied NEAR believed | | 84 |
| levied NEAR requested | | 293 |
| levied NEAR correlated | | 6 |

| Choice | | $\text{Score}_2$ |
|---|---|---|
| p(levied \| imposed) | 3,593 / 1,198,495 | 0.0029979 |
| p(levied \| believed) | 84 / 2,537,348 | 0.0000331 |
| p(levied \| requested) | 293 / 4,774,446 | 0.0000614 |
| p(levied \| correlated) | 6 / 244,353 | 0.0000246 |

## 3.3 Kea: Query Feature Set

The assumption behind the *keyphrase-frequency* feature is that documents in the same domain will tend to share keyphrases. In other words, the keyphrases in a given domain (e.g., computer science) will tend to be strongly *associated* with one another. The shared keyphrases tend to *co-occur* in the

*3. Removing Domain-Specificity: Mining for Lexical Knowledge*

domain. This suggests the following algorithm:

1. For a given document, use the baseline feature set to calculate the probability $p(key|T, D)$ of each candidate phrase. Make a list of the top $K$ candidates that have the highest probability of being keyphrases. (In the following experiments, $K$ is 4.) Make another list of the top $N$ candidates that will be further evaluated using the new query features. (We assume $N > K$. In the following experiments, $N$ is 100. Usually there are many more than 100 candidate phrases, but, for efficiency reasons, we do not re-evaluate all of the candidates. The list of the top $N$ candidates includes the top $K$ candidates.)
2. For each of the top $N$ baseline phrases, use PMI-IR to measure the strength of association with each of the top $K$ baseline phrases. These association strengths will be the new features, the *query* feature set.
3. For each of the top $N$ baseline phrases, use the new, extended feature set to revise the estimated probability that the candidate phrase is a keyphrase. If the user has requested $M$ phrases, then output the top $M$ phrases, according to the revised probability estimate.

The idea is that the top $K$ baseline phrases are the phrases that are most likely to be true keyphrases. Therefore, if a candidate phrase is strongly associated with one or more of the top $K$ baseline phrases, then it is more likely to be a keyphrase.

This is a two-pass algorithm. In the first pass, we run Kea with the baseline feature set. In the second pass, we run Kea with the query feature set. Kea is trained twice with the same training corpus. First it is trained with the baseline feature set, then, after the new features have been calculated, it is trained again with the query feature set.

Table 3 lists the twelve features in the query feature set that are calculated in step 2 above. All of the features are numerical. The first four features come from the baseline model. The first two features ($TF \times IDF$ and *distance*) are directly copied from the baseline feature set. The next two features (*baseline_rank* and *baseline_probability*) are calculated using the baseline model. The four features *rank_b1ci, ..., rank_b4ci* are based on equation (7). For example, *rank_b1ci* is the score for the $i$-th candidate phrase, $candidate_i$, calculated using equation (7) to measure the strength of association with the top-ranked baseline phrase, $baseline_1$. The four features *rank_b1capci*, ..., *rank_b4capci* are based on equation (6). In these latter four features, the $i$-th candidate phrase, $candidate_i$, is transformed to a new phrase, $cap\_candidate_i$, by converting the first character of each word in $candidate_i$ to upper case. According to the AltaVista® query syntax, the lower case query $candidate_i$ can match documents that contain the phrase $candidate_i$ in any combination of upper and lower case, but the capitalized query $cap\_candidate_i$ can only match documents that contain the phrase $candidate_i$ with the same capitalization as $cap\_candidate_i$. The hope is that the query $cap\_candidate_i$ will only retrieve documents in which the phrase $candidate_i$ appears in a title or section heading. Thus the four features *rank_b1capci*, ..., *rank_b4capci* are intended to measure the strength of association between the candidate phrases and the top baseline phrases when the candidate phrases appear in titles or section headings. I use equation (6) (based on "AND") with the latter four features because the capitalization is likely to reduce the number of hits. On the other hand, I use equation (7) (based on "NEAR") for the former four features because there are likely to be enough hits.

Features 5 to 12 are normalized as follows. Suppose that $\langle 0.8, 0.1, 0.3, 0.1 \rangle$ is a list of raw values for one feature. (That is, this is a column from the data table, not a row. There would usually be 100 raw values in this list, rather than just four.) First, the raw values are converted to ranks, where duplicate raw values map to the same rank. Thus we have $\langle 1, 3, 2, 3 \rangle$, if the raw values are sorted in descending order. Finally, the ranks are linearly normalized to range from 0 to 1, so we have

Table 3: The query feature set.

| | Name of feature | Description of feature |
|---|---|---|
| 1 | $TF \times IDF$ | Exactly the same as the baseline $TF \times IDF$ feature |
| 2 | *distance* | Exactly the same as the baseline *distance* feature |
| 3 | *baseline_rank* | The rank of the candidate phrase in the list of the top $N$ baseline keyphrases |
| 4 | *baseline_probability* | The baseline probability estimate $p(key \mid T, D)$ |
| 5 | *rank_b1ci* | The normalized rank of the candidate phrase $candidate_i$ when sorted by hits($baseline_1$ NEAR $candidate_i$) / hits($candidate_i$) |
| 6 | *rank_b2ci* | The normalized rank of the candidate phrase $candidate_i$ when sorted by hits($baseline_2$ NEAR $candidate_i$) / hits($candidate_i$) |
| 7 | *rank_b3ci* | The normalized rank of the candidate phrase $candidate_i$ when sorted by hits($baseline_3$ NEAR $candidate_i$) / hits($candidate_i$) |
| 8 | *rank_b4ci* | The normalized rank of the candidate phrase $candidate_i$ when sorted by hits($baseline_4$ NEAR $candidate_i$) / hits($candidate_i$) |
| 9 | *rank_b1capci* | The normalized rank of the candidate phrase $candidate_i$ when sorted by hits($baseline_1$ AND $cap\_candidate_i$) / hits($cap\_candidate_i$) |
| 10 | *rank_b2capci* | The normalized rank of the candidate phrase $candidate_i$ when sorted by hits($baseline_2$ AND $cap\_candidate_i$) / hits($cap\_candidate_i$) |
| 11 | *rank_b3capci* | The normalized rank of the candidate phrase $candidate_i$ when sorted by hits($baseline_3$ AND $cap\_candidate_i$) / hits($cap\_candidate_i$) |
| 12 | *rank_b4capci* | The normalized rank of the candidate phrase $candidate_i$ when sorted by hits($baseline_4$ AND $cap\_candidate_i$) / hits($cap\_candidate_i$) |

⟨1.0, 0.0, 0.5, 0.0⟩, if the first rank maps to 1 and the last rank maps to 0. The final normalization ensures that a feature with many duplicate values will span the same range as a feature with few duplicate values. Features are normalized per document, not per corpus. That is, I normalize the features one document at a time, without regard to any of the other documents. This is a local, contextual normalization, as opposed to a global normalization (Turney and Halasz, 1993). As with the other two feature sets, all of the query features are discretized (after normalization) using Fayyad and Irani's (1993) algorithm. (The discretization is global; it uses the whole training corpus.)

## 4. *Experiment 1: Comparison of Feature Sets on the CSTR Corpus*

This experiment compares the three feature sets using the setup of Frank *et al.* (1999). The same CSTR corpus is used, with the same training set of 130 documents and the same testing set of 500 documents. Kea 1.1.4 is used as the framework for comparing all three sets of features. The *keyphrase-frequency* feature is trained on a separate training set of 1,300 documents.[8] The baseline and query feature sets are trained using only the 130 training documents.

Figure 1 shows the experimental results. The desired number of output phrases varies from 1 to 20. For each requested number of output phrases, the plot shows the average number of output phrases that agree with the authors' phrases (the "correct" keyphrases). The plot shows that the keyphrase features perform best, followed by the query features, and lastly the baseline features.

---

8. I did not perform this training. Kea 1.1.4 is distributed with a pre-trained model for the *keyphrase-frequency* feature. The 1,300 training documents used to train *keyphrase-frequency* are from the CSTR corpus.

## 4. Experiment 1: Comparison of Feature Sets on the CSTR Corpus

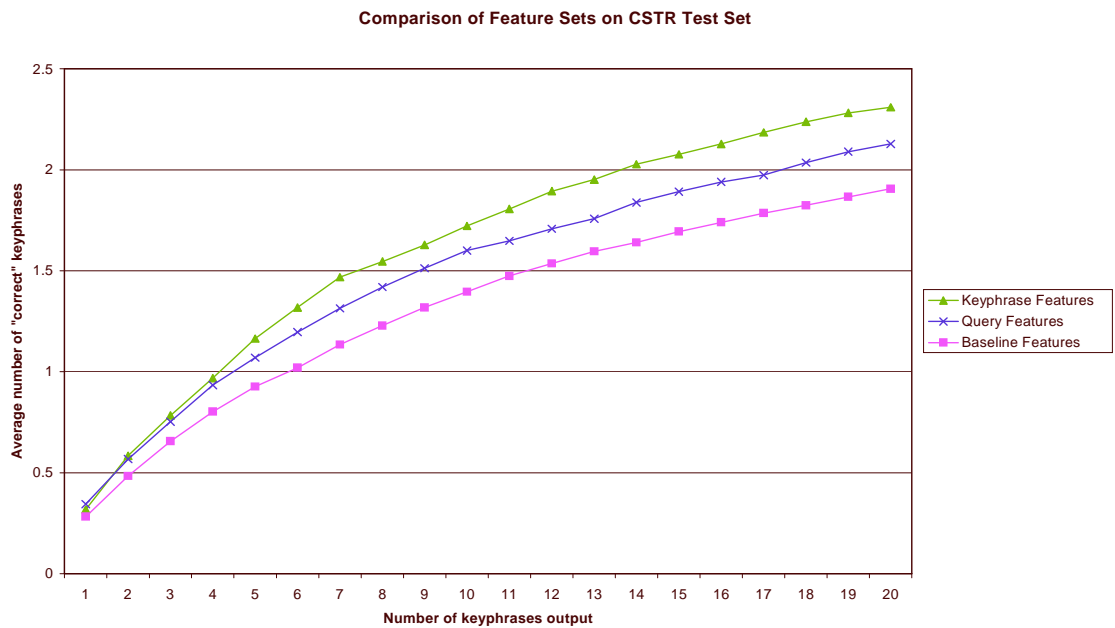

Figure 1: Comparison of the three feature sets on the CSTR corpus.

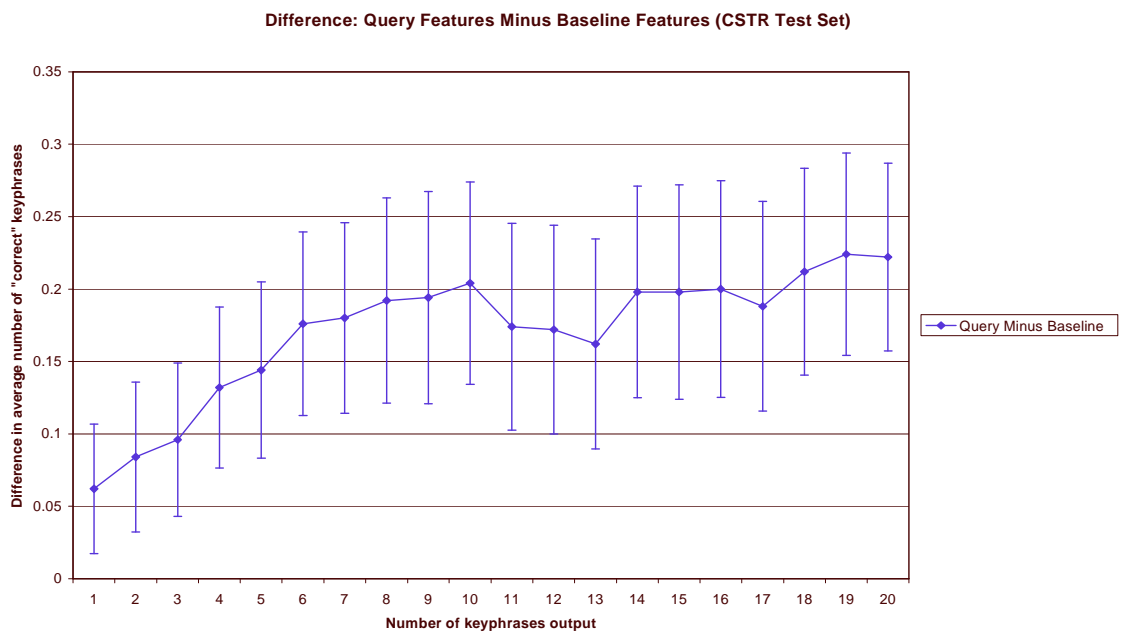

Figure 2: The difference between the query features and the baseline features.

*4. Experiment 1: Comparison of Feature Sets on the CSTR Corpus*

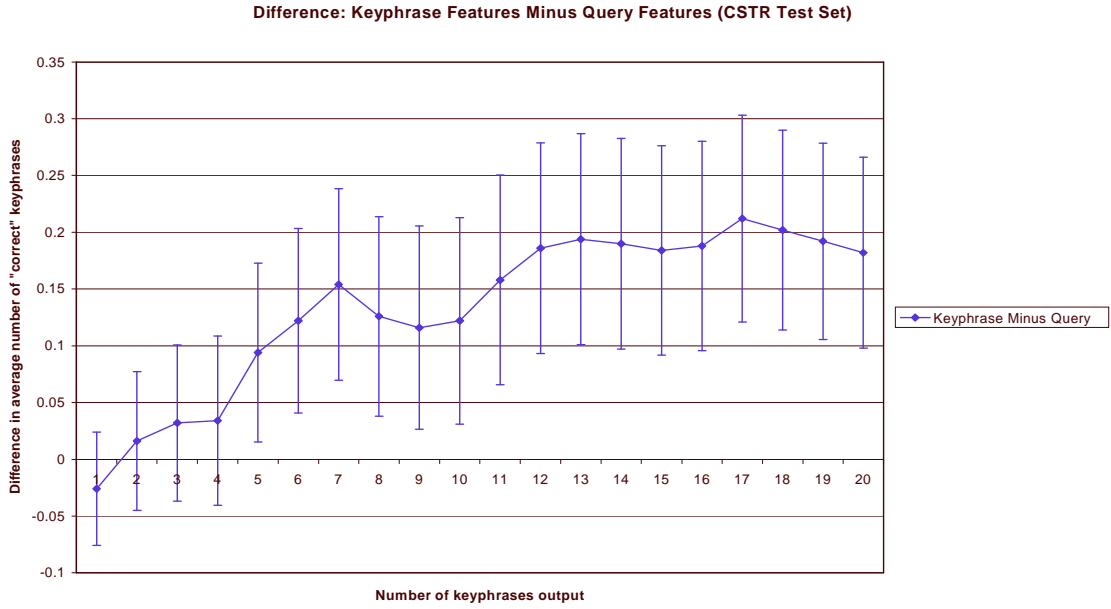

Figure 3: The difference between the keyphrase features and the query features.

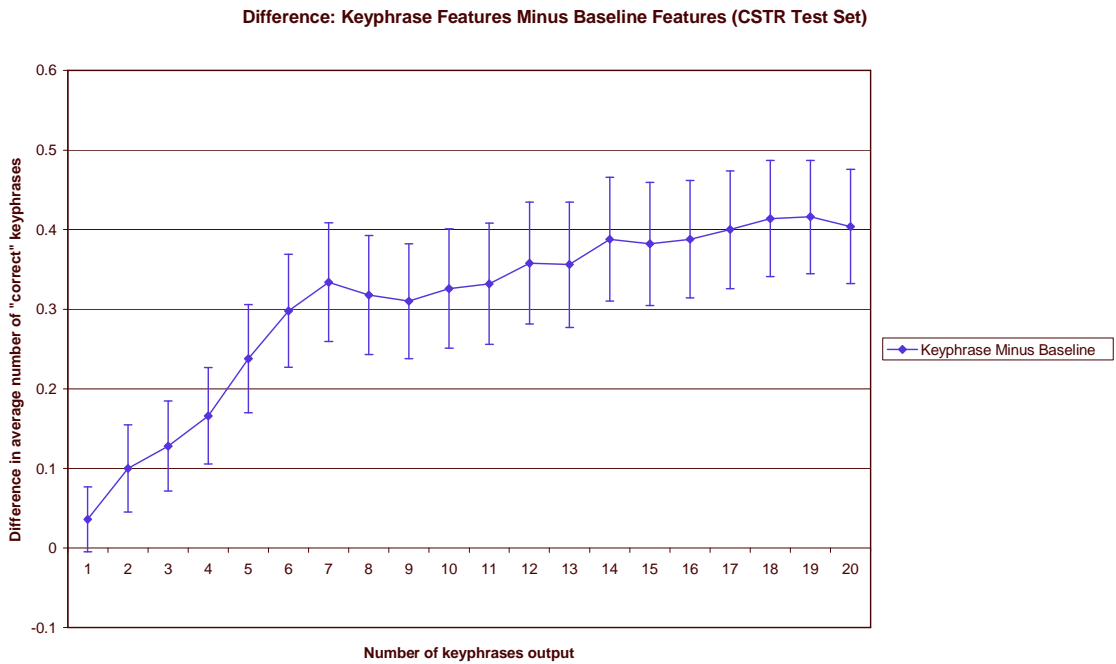

Figure 4: The difference between the keyphrase features and the baseline features.

I used the *paired t-test* to evaluate the statistical significance of the results (Feelders and Verkooijen, 1995). This is equivalent to applying the Student t-test to the differences between a pair of feature sets. Figure 2 looks at the differences between the query features and the baseline features. The error bars are 95% confidence regions. The performance of the query features is significantly better than the performance of the baseline features throughout the range of the desired number of output phrases. Figure 3 displays the difference between the keyphrase features and the query features. The performance of the keyphrase features is significantly better than the performance of the query features when five or more phrases are desired. Finally, Figure 4 plots the difference between the keyphrase features and the baseline features. The performance of the keyphrase features is significantly better except when only one phrase is output.

The experiment shows that the query feature set improves on the baseline feature set, although the improvement over the baseline is even larger with the keyphrase feature set. On the other hand, the query feature set does not require the additional 1,300 training documents that are used for the keyphrase feature set. However, the query feature set does use 350 million unlabeled documents.

Table 4 gives some examples of the output phrases for the three different feature sets. These are the top ten phrases for a document chosen randomly from the CSTR testing set, entitled, "Set-Based Bayesianism", by Kyburg and Pittarelli. Matches with the authors are italicized. (These examples are intended to give the reader an impression of the typical output of the algorithms. They are not intended to make any special point.)

Table 4: Examples of extracted keyphrases and the authors' keyphrases.

| Baseline features | Query features | Keyphrase features | Authors' keyphrases |
|---|---|---|---|
| probability functions | probability functions | convex | Bayesian methods |
| Bayesianism | Bayesianism | probability | decision-making |
| convex | Kyburg | *maximum entropy* | maximum entropy |
| strict Bayesian | convex | probability functions | uncertainty |
| convex Bayesianism | probabilities | agent | |
| Kyburg | set based Bayesianism | Bayesianism | |
| classical probability | set based | belief | |
| classical probability functions | expected utility | belief function | |
| *maximum entropy* | classical probability | probabilities | |
| agent | *maximum entropy* | strict Bayesian | |

## 5. *Experiment 2: Generalization from CSTR to LANL*

This experiment evaluates how well the learned models generalize from one domain to another. The training domain is the CSTR corpus, using exactly the same training setup as in the first experiment. The testing domain consists of 580 documents from the LANL collection (physics papers from the arXiv repository at the Los Alamos National Laboratory).[9] This corpus consists of PostScript files that have been submitted to the arXiv repository by physicists around the world. The documents are journal papers, conference papers, technical reports, and preprints in physics. There is a fair amount of noise in the corpus, because the PostScript files were automatically converted to plain text. For each document in the corpus, the author's keyphrase list was removed from the title page and placed

---

9. See http://www.arxiv.org/.

in a separate file.

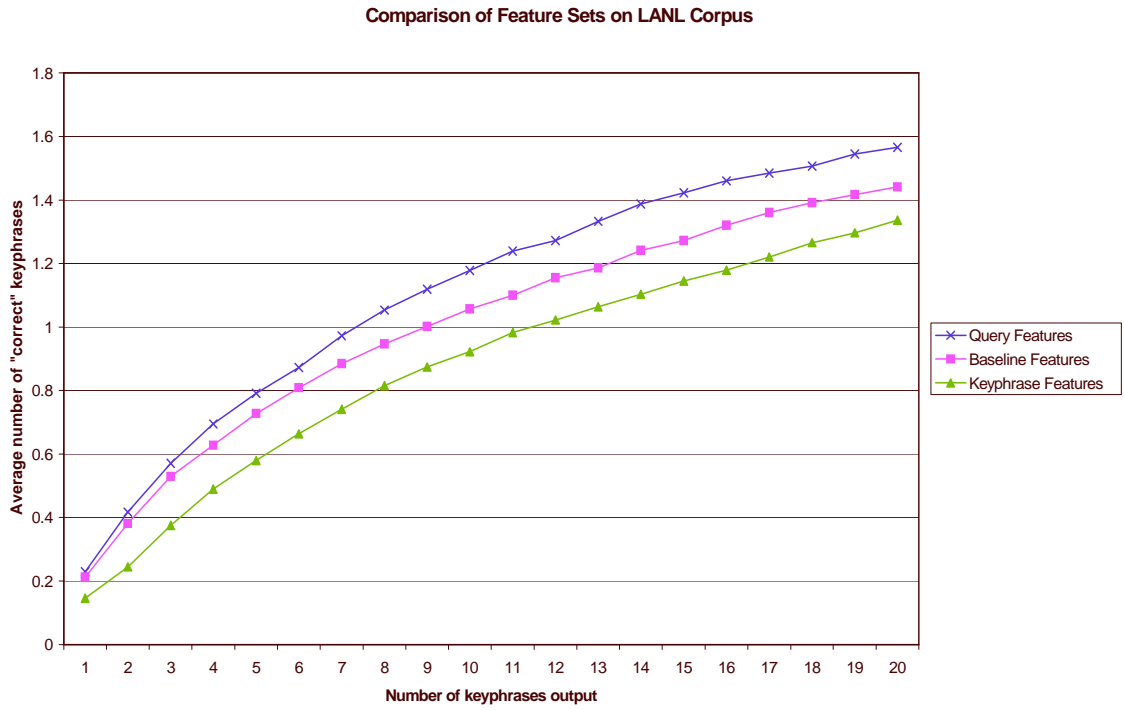

Figure 5: Comparison of the three feature sets on the LANL corpus.

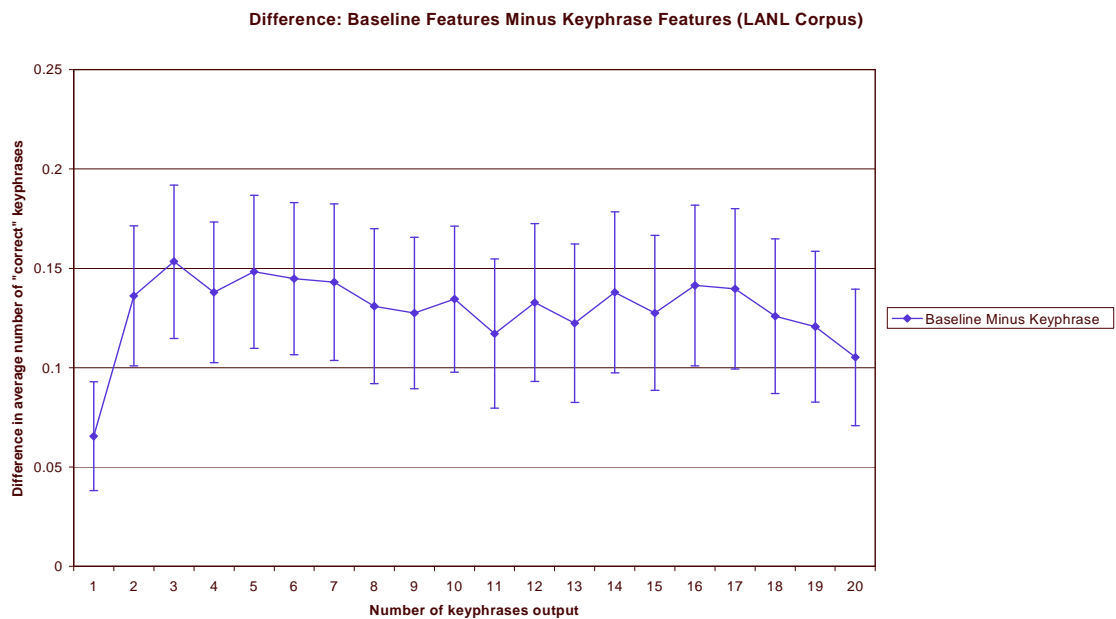

Figure 6: The difference between the baseline features and the keyphrase features.

## 5. Experiment 2: Generalization from CSTR to LANL

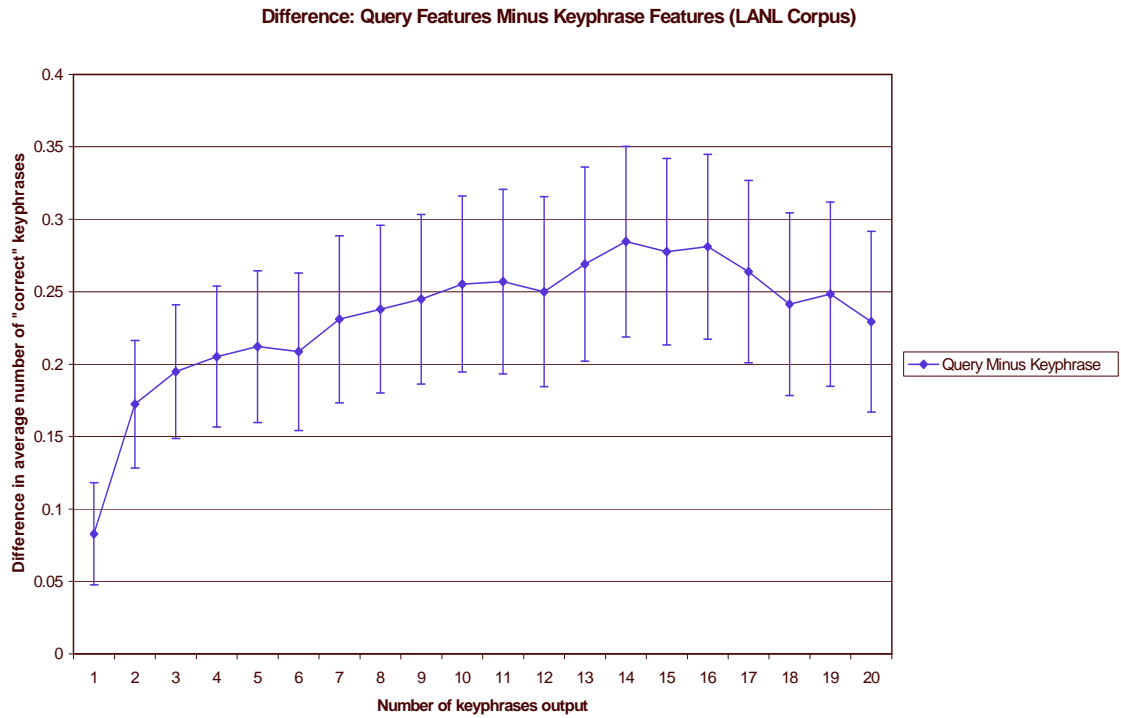

Figure 7: The difference between the query features and the keyphrase features.

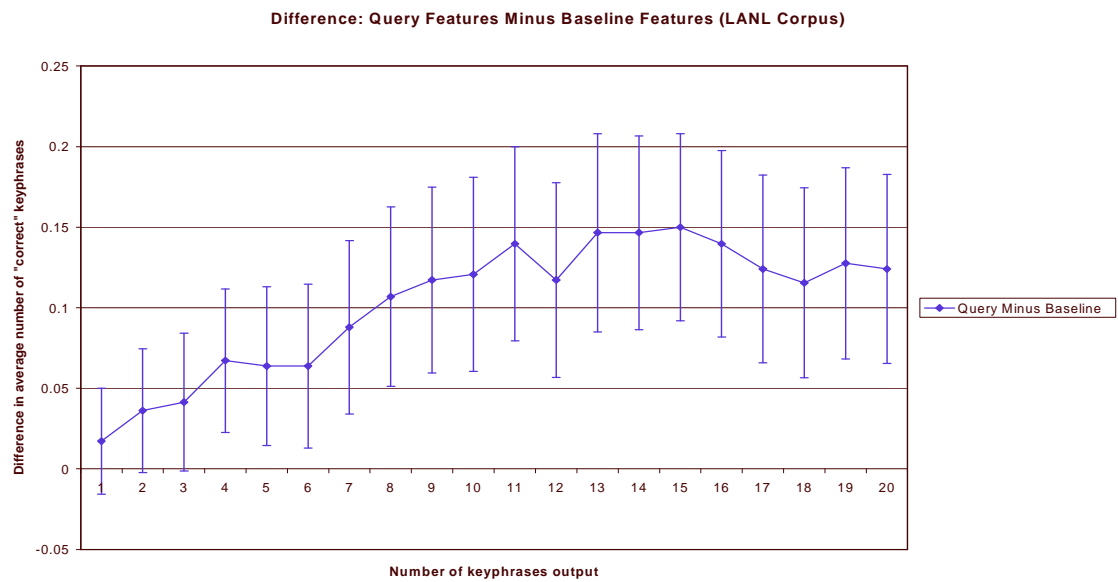

Figure 8: The difference between the query features and the baseline features.

Figure 5 shows that the query feature set has the best performance, followed by the baseline features, and lastly the keyphrase features. Paired t-tests show that the differences are statistically significant at the 95% confidence level. Figure 6 demonstrates that the performance of the baseline features is significantly better than the performance of the keyphrase features for all numbers of desired phrases, from 1 to 20. Figure 7 illustrates that the query features are significantly better than the keyphrase features. Finally, Figure 8 shows that the query features are significantly better than the baseline features when more than three phrases are output. For less than three phrases, the query features are still better than the baseline features, but the difference is not significant at the 95% confidence level.

The experiment illustrates the domain-specificity of the keyphrase features, which perform worse than the baseline features in this situation. The experiment also shows that the query features are not domain-specific; they generalize well from the CSTR corpus to the LANL corpus.

## 6. *Experiment 3: Evaluation of Feature Subsets*

There are twelve features in the query feature set. This experiment looks at various subsets of the twelve features, to see whether all of them are required. Table 5 lists the feature subsets that are compared in this experiment. Figure 9 plots the results of the experiment with the CSTR corpus. Table 6 summarizes the results of the paired t-tests. In general, it appears that all of the features are useful, but the results are not always statistically significant.

Table 5: Feature subsets that are evaluated in Experiment 3.

| Feature subsets | Description of subset |
| --- | --- |
| Group 0 | All twelve features |
| Group 1 | All except *baseline_rank* and *baseline_probability* |
| Group 2 | All except *TF × IDF* and *distance* |
| Group 3 | All except *rank_b1ci*, …, *rank_b4ci* |
| Group 4 | All except *rank_b1capci*, …, *rank_b4capci* |

Table 6: Results of the paired t-tests for the feature subsets on the CSTR testing set.

| Pairs of groups | Summary of paired t-test results |
| --- | --- |
| Group 1 - Group 0 | Group 0 is significantly better, except when 3 phrases are output, in which case there is no significant difference. |
| Group 2 - Group 0 | Group 0 is better, except when 17 phrases are output, but the differences are mostly not significant. |
| Group 3 - Group 0 | Group 0 is better, except when 3 phrases are output, but the differences are mostly not significant. |
| Group 4 - Group 0 | Group 0 is better for all numbers of output phrases, and the differences are usually significant. |

## 6. Experiment 3: Evaluation of Feature Subsets

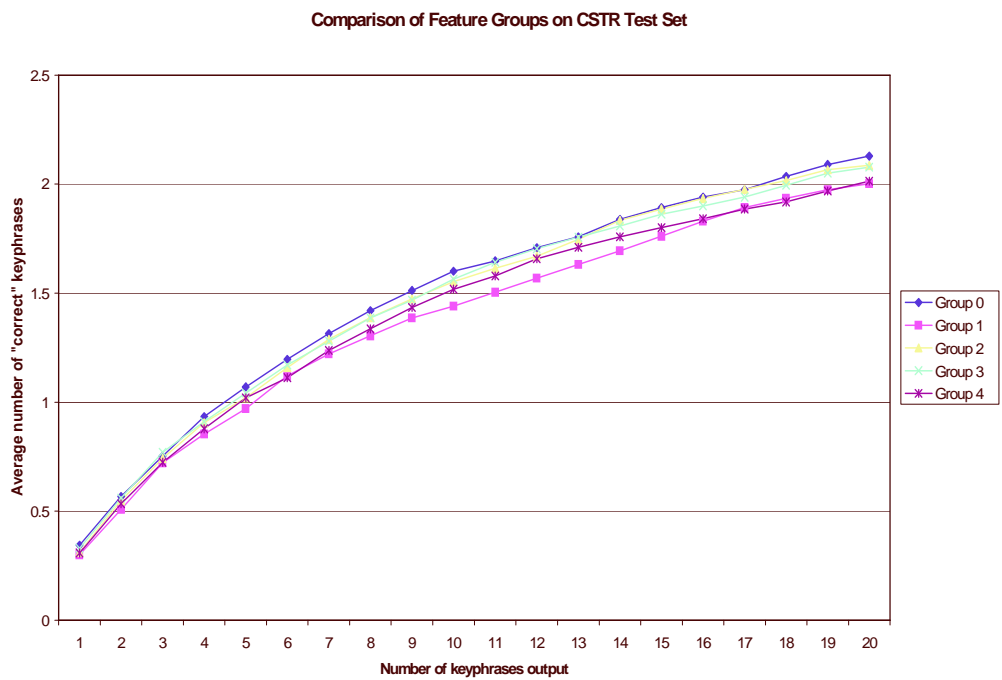

Figure 9: A comparison of various feature subsets on the CSTR testing set.

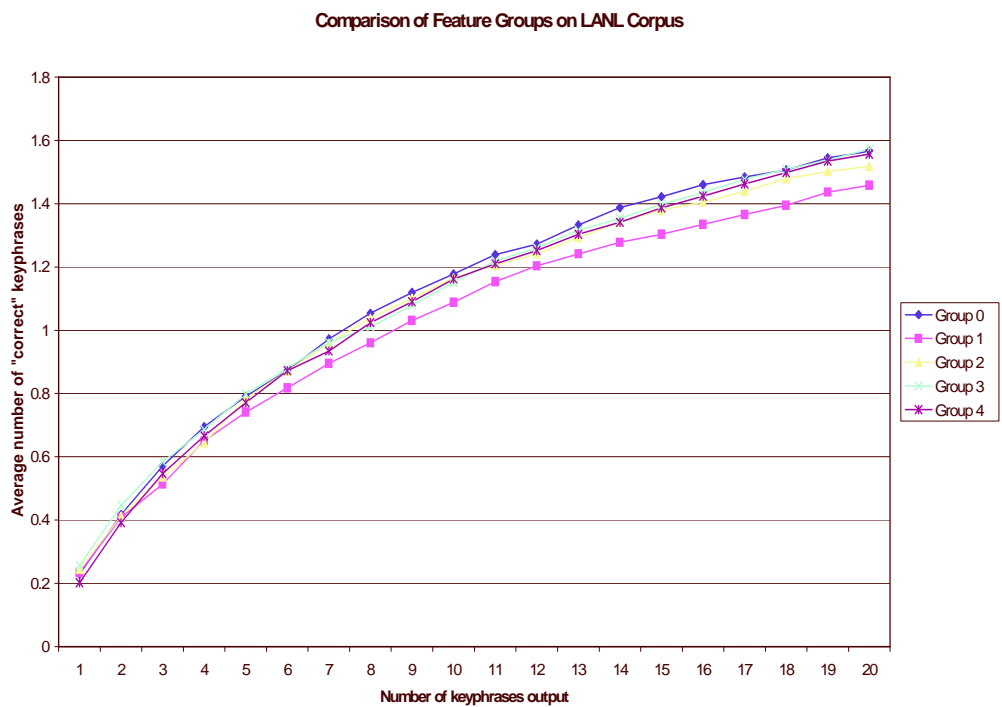

Figure 10: A comparison of various feature subsets on the LANL corpus.

Figure 10 evaluates the same feature subsets on the LANL corpus. The experimental setup is like the setup in Experiment 2: the models are trained on the CSTR training data and then tested on the LANL corpus. Table 7 summarizes the results of the paired t-tests. As with the CSTR testing set, it appears that all of the features are useful, but the results are not always statistically significant.

Table 7: Results of the paired t-tests for the feature subsets on the LANL corpus.

| Pairs of groups | Summary of paired t-test results |
| --- | --- |
| Group 1 - Group 0 | Group 0 is significantly better, except when 1 or 2 phrases are output, in which case there is no significant difference. |
| Group 2 - Group 0 | Group 0 is better, except when 1 phrase is output, but the differences are not always significant. |
| Group 3 - Group 0 | Group 0 is better more often than it is worse, but the differences are mostly not significant. |
| Group 4 - Group 0 | Group 0 is never worse, but the differences are mostly not significant. |

## 7. *Experiment 4: Relations Among Feature Sets*

The *keyphrase-frequency* feature and the query features are based on the same basic concept: the keyphrases in a given domain will tend to be strongly associated with one another. Therefore I conjectured that there would be a large similarity between the output phrases with the keyphrase feature set and the output phrases with the query feature set, when the models are trained on the CSTR training set and tested on the CSTR testing set (as in Experiment 1). This experiment tests this conjecture.

Figure 11 presents the overlap among the phrases that are extracted with the three different feature sets. This figure focuses on the output phrases that overlap with the authors' keyphrases. The desired number of output phrases is set to 20. If we look at the bar for the baseline feature set, we see that the total height of the bar is 1.9, meaning that, when the desired number of output phrases is 20, on average 1.9 of the 20 keyphrases will match the authors' keyphrases. The part of this bar that is labeled "Shared with All" has a height of 1.3, which means that, on average 1.3 of these phrases will be shared with the phrases that are output by both the keyphrase and query feature sets. The part of this bar that is labeled "Shared with Query" represents the keyphrases that are output by both the baseline feature set and the query feature set, but not the keyphrase feature set. "Shared with Keyphrase" represents the keyphrases that are output by both the baseline feature set and the keyphrase feature set, but not the query feature set. "Shared with Baseline" represents the keyphrases that are output by the baseline feature set only.

Figure 11 shows that the baseline and query feature sets have many keyphrases in common, but the output of the keyphrase feature set is quite different from the output of the other two feature sets. This is evidence against the conjecture that the query and keyphrase feature sets would produce similar output.

An objection to this analysis is that it only looks at the output phrases that match with the authors' keyphrases. To answer this objection, Figure 12 shows the overlap in the output phrases when all 20 output phrases are considered, without regard to whether they match with the authors' keyphrases. This figure shows the same general pattern as the previous figure: the baseline and query feature sets have many phrases in common, but the output of the keyphrase feature set is quite different. Therefore it appears that the conjecture is false. In spite of their conceptual similarity, the query features behave quite differently from the *keyphrase-frequency* feature.

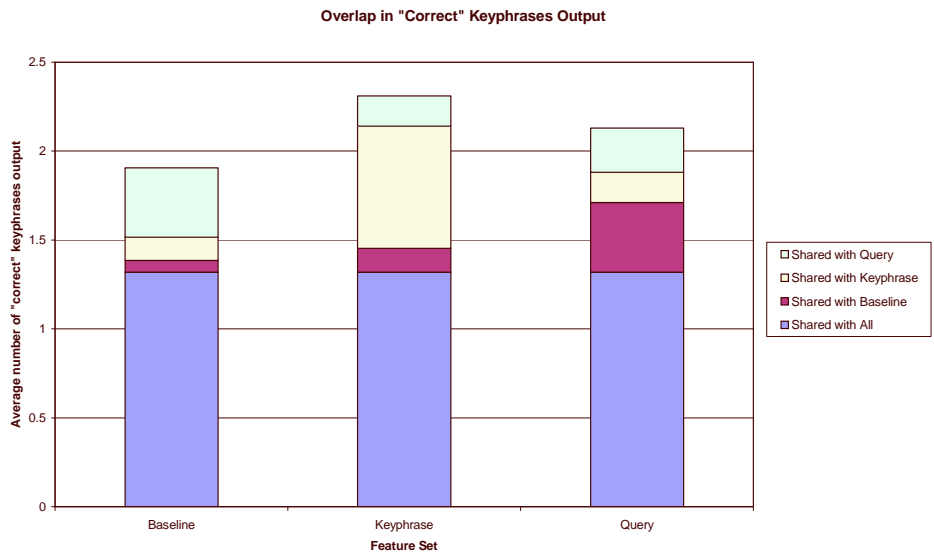

Figure 11: The overlap among the feature sets in the output phrases that match with the authors' phrases.

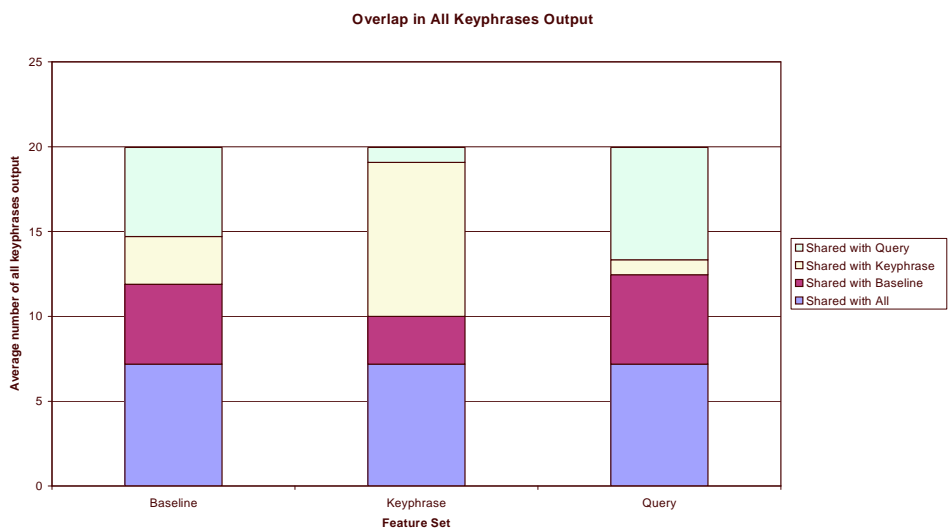

Figure 12: The overlap among the feature sets in the output phrases, without regard to whether the phrases agree with the authors' keyphrases.

## 8.  *Experiment 5: Combining Query and Keyphrase Features*

Experiment 4 suggests that the query features and the *keyphrase-frequency* feature are independent, so there may be some value to combining them. Experiment 5 evaluates such a hybrid feature set. This requires a small modification to the algorithm described in Section 3.3. We still have a two-pass algorithm, but the first pass now uses the keyphrase feature set instead of the baseline feature set.

## 8. Experiment 5: Combining Query and Keyphrase Features

The second pass now uses thirteen features, the twelve query features plus the *keyphrase-frequency* feature. The *baseline_rank* and *baseline_probability* features become *keyphrase_rank* and *keyphrase_probability*.

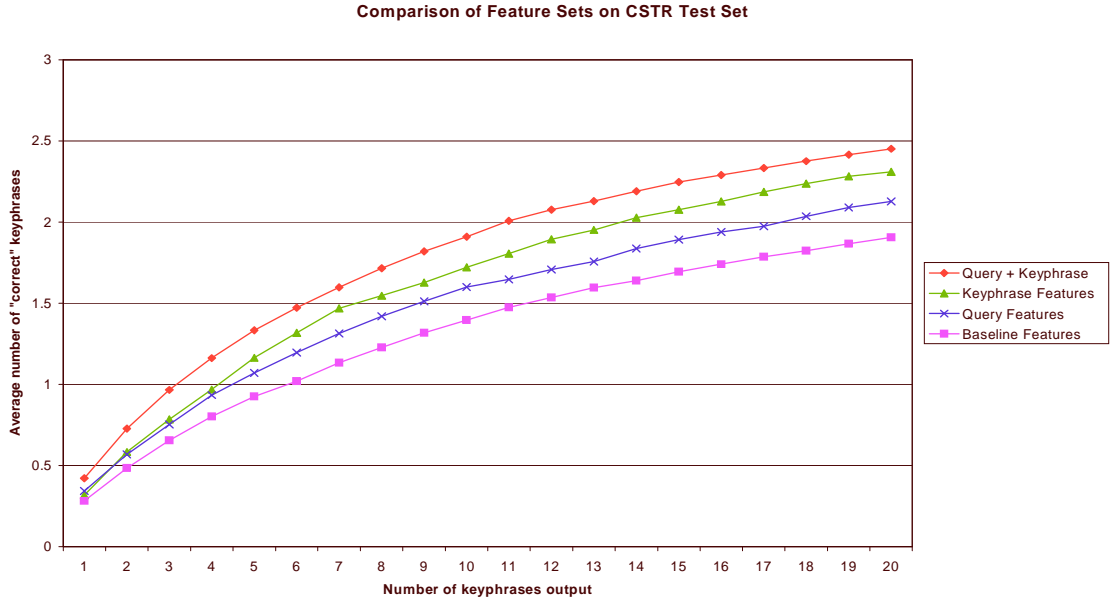

Figure 13: Comparison of a hybrid of the query and keyphrase feature sets with the other three feature sets on the CSTR corpus.

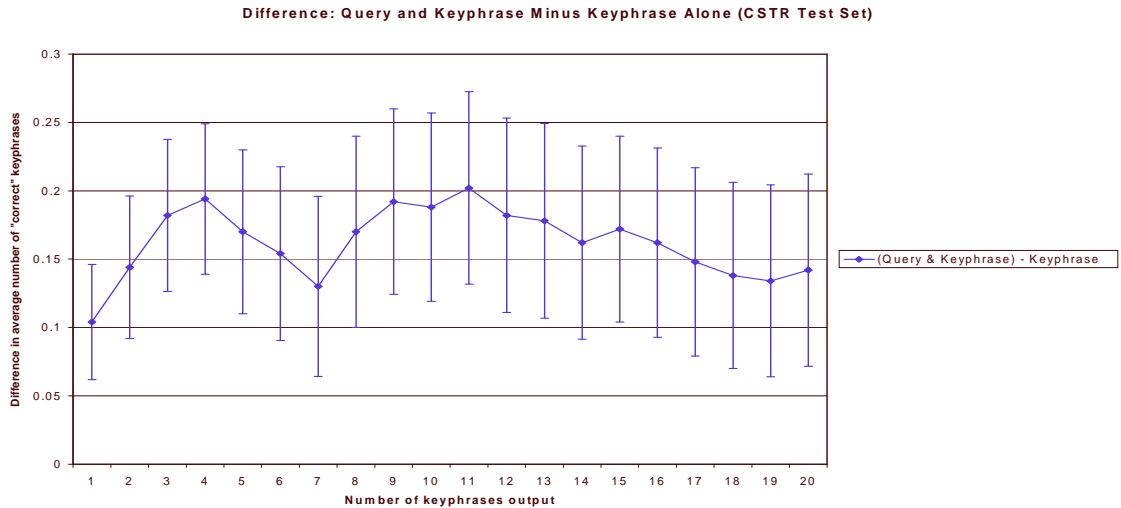

Figure 14: Paired t-test of the hybrid feature set versus the keyphrase feature set on the CSTR corpus.

## 8. Experiment 5: Combining Query and Keyphrase Features

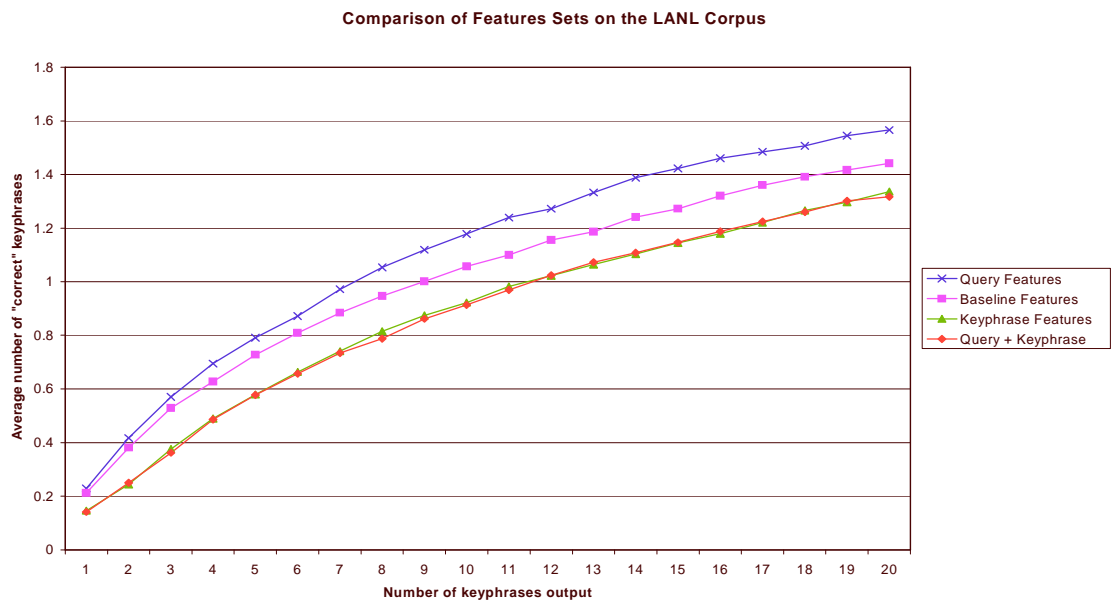

Figure 15: Comparison of a hybrid of the query and keyphrase feature sets with the other three feature sets on the LANL corpus.

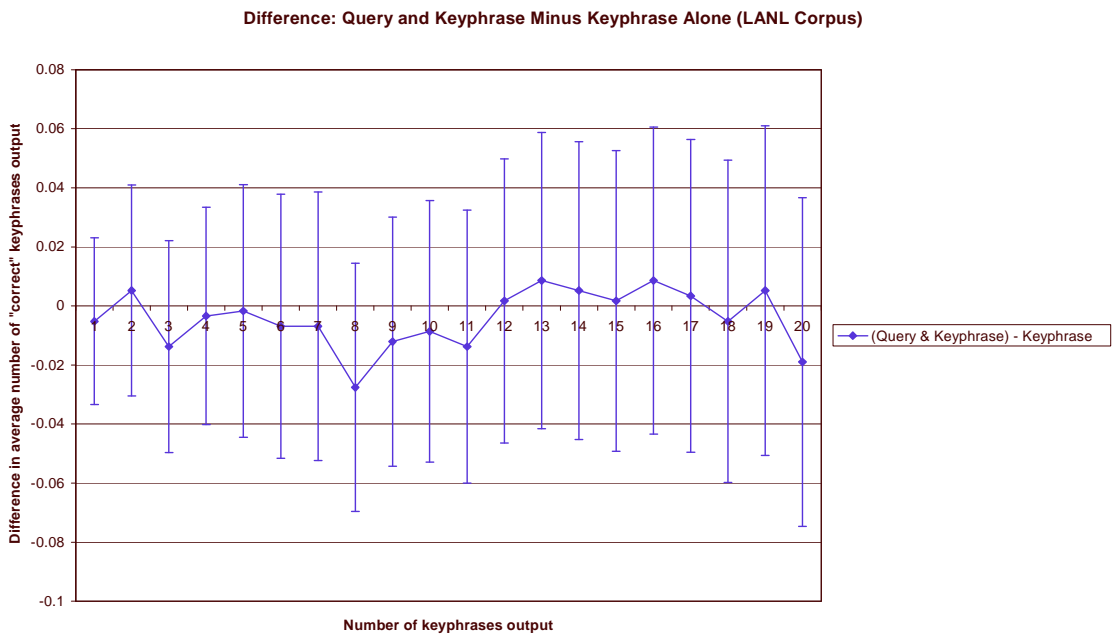

Figure 16: Paired t-test of the hybrid feature set versus the keyphrase feature set on the LANL corpus.

Figure 13 plots the results when the hybrid feature set is trained and tested on the CSTR data. Figure 14 shows that the hybrid feature set performs significantly better than the closest competitor, the keyphrase feature set.

Figure 15 shows the results when the hybrid feature set is trained on the CSTR corpus and then evaluated on the LANL corpus. In this case, as we can see from Figure 16, there is no significant difference between the hybrid features and the keyphrase features alone.

The experiment shows that, when the appropriate training data are available, it can be beneficial to combine the query feature set and the keyphrase feature set. When the data are not available, it is best to use the query feature set alone. Thus the hybrid feature set has the same limitations as the keyphrase feature set: domain-dependency and training-intensiveness.

## 9. *Experiment 6: Evaluating Keyphrases by Familiarity*

In the experiments so far, I have measured the performance of the various feature sets by the level of agreement between the authors and the machine. As I mentioned in Section 2.1, this underestimates the quality of the machine-extracted phrases. One way to get a better estimate is to ask human readers to evaluate the machine-extracted phrases (Turney, 2000), but this is a laborious exercise. In this experiment and the next experiment, I examine alternative ways to automatically evaluate keyphrases. The intent is not to replace the author as a standard of performance; the motivation is to use a variety of different performance measures to gain more insight into automatic keyphrase extraction.

The aim of the current experiment is to evaluate whether the machine-extracted phrases seem reasonable or familiar for documents in the given domain, without regard to whether they are appropriate for the given input document. In this experiment, the performance of the feature sets will be measured by the level of agreement between the machine-extracted phrases for a given document and the author-assigned phrases for any document in the testing set, including the given input document. Of course, this performance measure is heavily biased in favour of the keyphrase feature set, at least when the testing domain corresponds to the training domain. However, as long as we are aware of this bias, it may still be interesting to see what happens.

Figure 17 presents the experimental results when the features are trained and tested with the CSTR corpus (as in Experiment 1). The keyphrase features perform best, as expected. The query and baseline features perform about the same. The paired t-test shows that the difference between the keyphrase features and the other features is significant, but the difference between the query features and the baseline features is mostly insignificant.

Figure 18 shows the results when the features are trained on the CSTR corpus and then tested with the LANL corpus (as in Experiment 2). The paired t-test shows that the differences among the feature sets become significant when seven or more phrases are output, at which point the query features perform best, followed by the keyphrase features, and lastly the baseline features.

It is interesting to contrast Figure 18 in this experiment with Figure 5 in Experiment 2, where the keyphrase features performed better than the baseline features. This suggests that, although the phrases extracted by the keyphrase features from the LANL corpus may not be appropriate for the documents (as indicated by Figure 5), they are still reasonable for the domain (as indicated by Figure 18). This result (Figure 18) is somewhat surprising, since it was not expected that the keyphrase model trained on the CSTR corpus would produce reasonable phrases for the LANL corpus.

It is also interesting to contrast the performance of the query features in Figures 17 and 18. The phrases extracted by the query features seem to be much more reasonable for the LANL corpus than for the CSTR corpus. I have no explanation for this. It may be related to differing degrees of homo-

## 9. Experiment 6: Evaluating Keyphrases by Familiarity

geneity in the two domains or the two corpora.

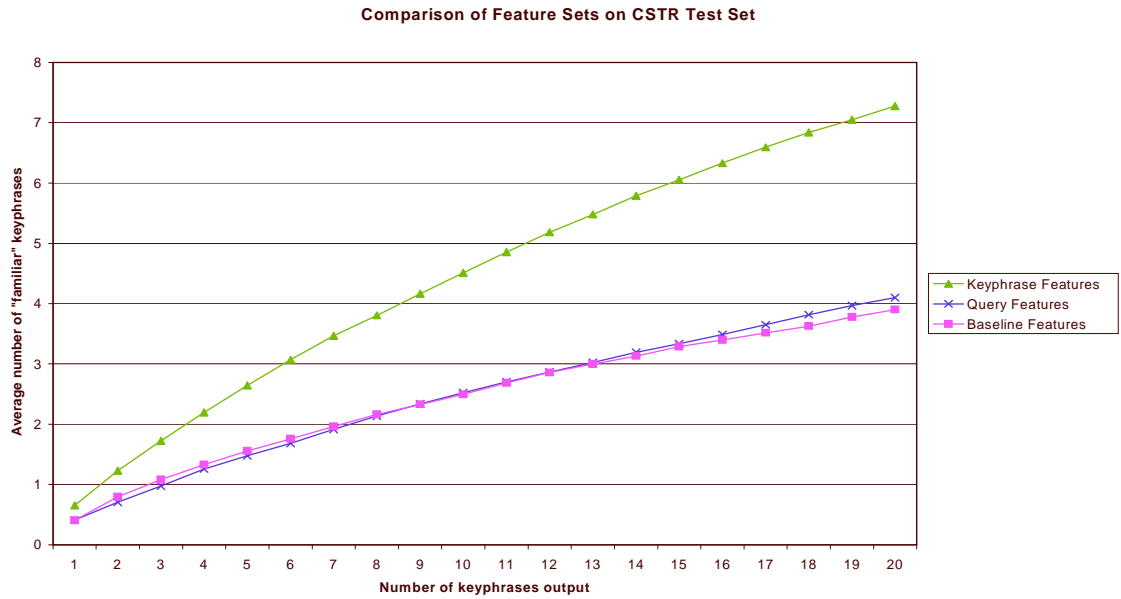

Figure 17: Comparison of the feature sets on the CSTR testing set, using familiarity as a performance measure.

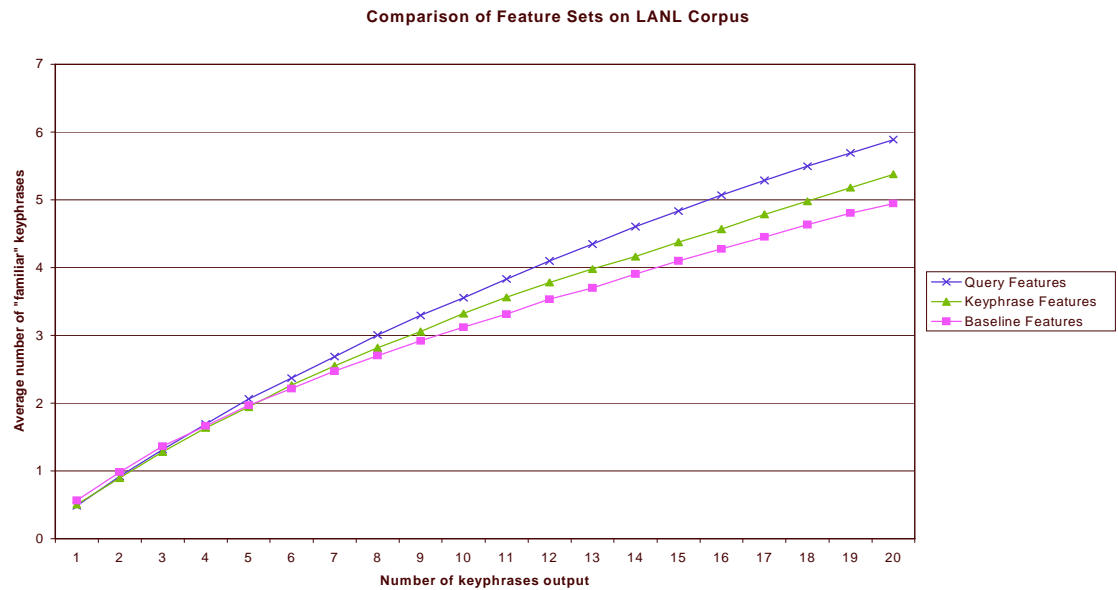

Figure 18: Comparison of the feature sets on the LANL Corpus, using familiarity as a performance measure.

## 10. *Experiment 7: Evaluating Keyphrases by Searching*

Martin and Holte (1998) argue that keyphrases make good query terms for finding documents that are similar or identical to a given document. This suggests a method for evaluating keyphrases. Good keyphrases should be specific enough that they can be used to find the original document, from which they were extracted, yet they should be general enough that they can also find many similar documents. It would be easy to make a query so specific that it only retrieved one particular document, or so general that it retrieved a huge number of documents. It is more difficult to make a query that balances specificity and generality. I believe that authors are faced with this same problem when they choose keyphrases for their documents. They want general keyphrases that will attract large audiences for their documents, yet they also want keyphrases that are specific enough to accurately capture the actual topics and contents of their documents.

For the CSTR corpus, I used the CSTR search engine, part of the New Zealand Digital Library at the University of Waikato.[10] This search engine indexes about 46,000 computer science papers.[11] The query syntax only allows searching by words (not phrases). The user can control whether the query matches documents that contain "all" or "some" of the words. When a query matches more than 50 documents, the search engine reports, "More than 50 documents matched the query," but does not give the actual number of hits. The search results are reported with 20 hits listed per page of results.

Figure 19 presents the results of experiments with the CSTR search engine. In this case, the feature sets have been trained and tested on the CSTR corpus, as in Experiment 1. For each of the 500 testing documents, a query was issued to the CSTR search engine, based on the top three output phrases for the various feature sets and the first three phrases for the author-assigned keyphrases. For a given set of three phrases, a query was generated by taking the conjunction of all of the words in the three phrases (an "all" query). The *specificity* of a query was measured by the probability that the top 20 hits included the original source document, from which the keyphrases were extracted. The *generality* of a query was measured by the probability that the query would return more than 50 documents. The bar chart includes error bars, which show the 95% confidence regions, calculated using the Student t-test (not a paired t-test this time). The chart shows that there are no significant differences among the phrases generated by the baseline features, the query features, and the authors, with respect to specificity and generality. However, the phrases generated by the keyphrase features are significantly different from the other three groups. They are both less specific (less likely to retrieve the source document in the top 20 hits) and more general (more likely to have more than 50 hits). This suggests that there is a systematic bias towards generality in the phrases that are selected with the keyphrase feature set.

For the LANL corpus, I used the CERN Document Server, a service of the European Organization for Nuclear Research.[12] This search engine indexes about 42,000 physics papers.[13] The query syntax allows searching by words or by phrases. The search engine returns the number of hits for each query. The user can set the number of hits listed per page of results. The choices are 10, 25, 100, or 500 hits per page. In this experiment, I chose 25 hits per page.

---

10. See http://www.nzdl.org/cgi-bin/cstrlibrary?a=p&p=about.
11. The main page reported 45,720 documents (http://www.nzdl.org/cgi-bin/cstrlibrary?a=p&p=about). I verified that all of the 500 CSTR testing set documents used in this paper were indexed by the search engine.
12. See http://weblib.cern.ch/fulltext.php.
13. The query "the" returned 41,765 hits. The arXiv site reported that there were 178,569 papers in the LANL archive (http://arxiv.org/show_monthly_submissions), but it appears that they were not all indexed by the CERN document server. I verified that all of the 580 papers in the LANL corpus used in this paper were indexed by the search engine.

## 10. Experiment 7: Evaluating Keyphrases by Searching

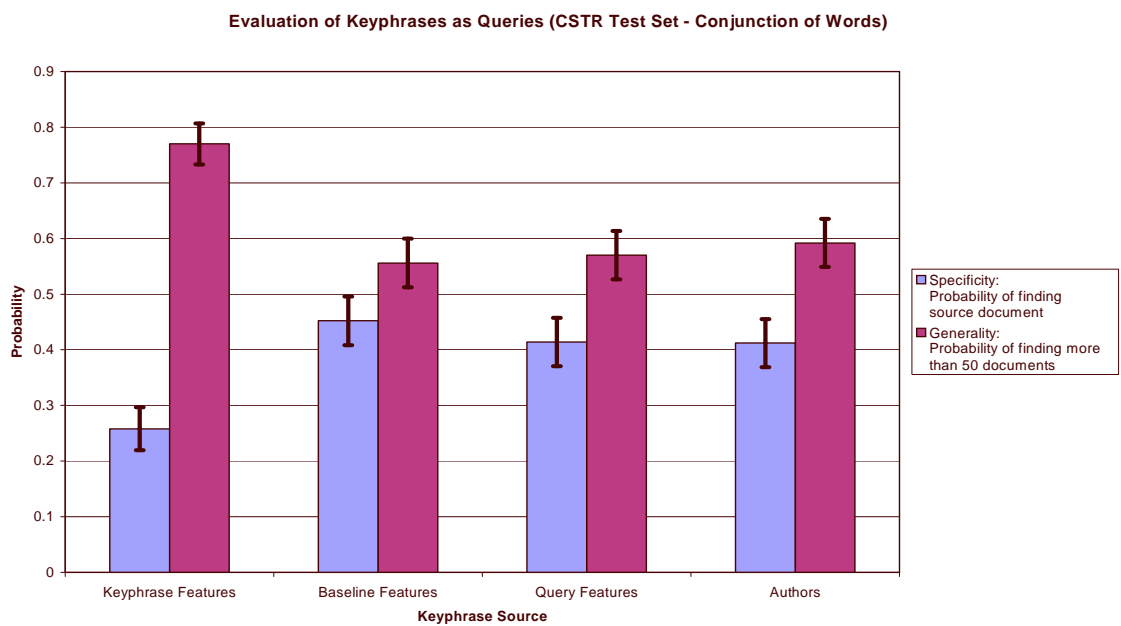

Figure 19: Comparison of the feature sets and the authors' keyphrases on the CSTR corpus, using searching as a performance measure.

Figure 20 displays the results with the LANL search engine. The feature sets were trained with the CSTR corpus and then tested with the LANL corpus, as in Experiment 2. For each of the 580 testing documents, a query was issued to the LANL search engine, based on the top three output phrases for the various feature sets and the first three phrases for the author-assigned keyphrases. For a given set of three phrases, a query was generated by taking the conjunction of the phrases (not the individual words, unless the phrases were only one-word long). The *specificity* of a query was measured by the probability that the top 25 hits (not the top 20) included the original source document, from which the keyphrases were extracted. The *generality* of a query was measured by the probability that the query would return more than 50 documents. The bar chart shows that there are no significant differences between the baseline and query features, in terms of generality and specificity. However, the authors' keyphrases are significantly less specific *and* less general than the baseline and query phrases. Again, the phrases selected by the keyphrase feature set are significantly more general than the phrases selected by the other three approaches, although the specificity is about the same. This supports the hypothesis that the keyphrase feature set has a bias towards generality.

One difference between Figures 19 and 20 is that the former queries were words and the latter queries were phrases. The CSTR search engine does not handle phrases, but the LANL search engine does handle words. Therefore I repeated the LANL experiment with words instead of phrases. As we can see in Figure 21, the results follow the same general pattern. As expected, the word-based queries are more general and less specific than the phrase-based queries (compare Figures 20 and 21).

*Turney* 29

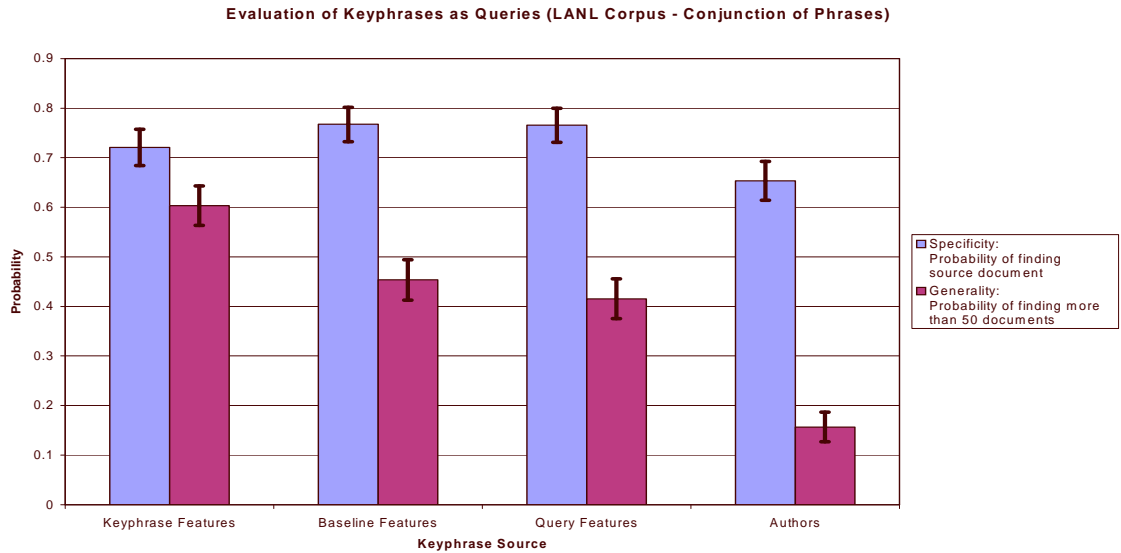

Figure 20: Comparison of the feature sets and the authors' keyphrases on the LANL corpus, using searching as a performance measure (phrase-based queries).

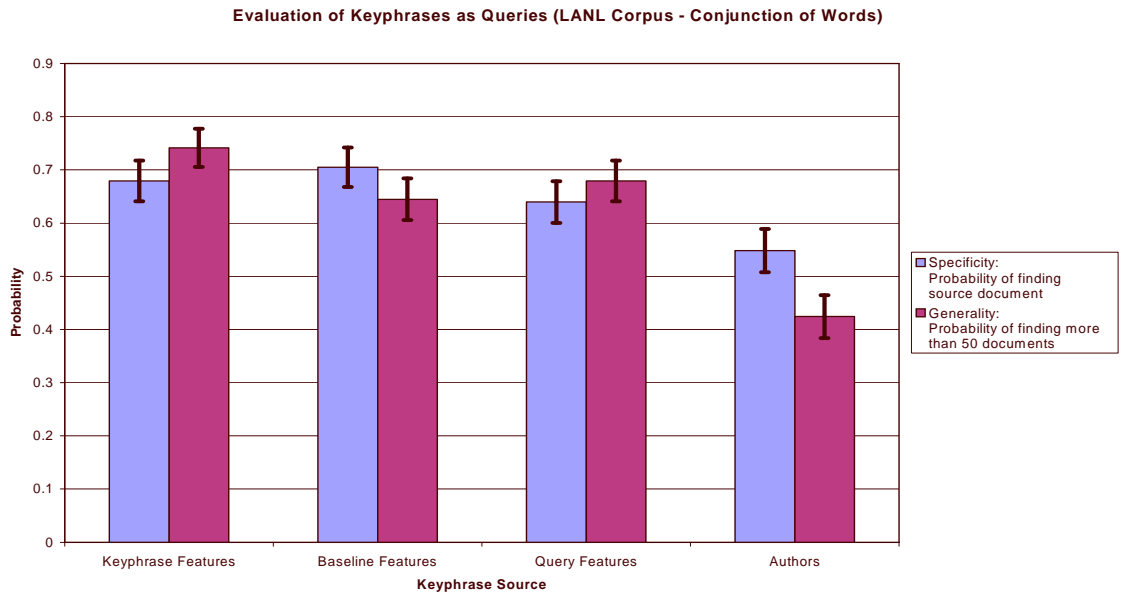

Figure 21: Comparison of the feature sets and the authors' keyphrases on the LANL corpus, using searching as a performance measure (word-based queries)

## 11.   *Discussion: Limitations, Applications, Future Work*

The main limitation of the new query feature set is the time required to calculate the query-based features. Examination of Table 3 shows that it takes 10 queries to AltaVista to calculate one feature vector. Each query takes about one second. At 10 queries per feature vector times 1 second per query times 100 feature vectors per document, we have 1,000 seconds per document (roughly 15 minutes per document). The time required by the other aspects of Kea is completely insignificant compared to this. I accelerated the process a small amount by caching query results. It could be further accelerated by using multi-threading to issue queries in parallel, although this may not be considered ethical by the provider of the search engine service.[14]

The vast majority of the time required is taken up by network traffic between the computer that hosts the search engine and the computer that hosts the software that is calculating the query features. With a local search engine, the time required could be reduced to less than one second per document. If we extrapolate the current trends in hardware, then it seems that the typical desktop personal computer will be able to locally index and search 350 million web pages easily within about ten years. Thus, although the new query features are currently impractical for many types of applications, it will not be long before this in no longer an issue. Once the problem of speed has been solved, the query features will be useful to improve the quality of keyphrase extraction in all of the various applications for keyphrases (summarizing, indexing, labeling, categorizing, clustering, highlighting, browsing, searching, etc.).

There are many questions that remain to be addressed by future research: Does a significant improvement in performance, as measured by agreement with the author, result in significant improvement in performance in the various applications for keyphrases? How does the improvement vary as the size of the unlabeled corpus (the 350 million Web pages) varies? Could good results be obtained using a much smaller corpus? Perhaps the CSTR search engine would be sufficient to generate query features for the CSTR corpus, which would result in features that are domain-specific, but not training-intensive. Similarly, the LANL search engine might be sufficient for features for the LANL corpus. What other techniques could be used to exploit unlabeled data for improved keyphrase extraction? What other features could improve keyphrase extraction? What biases are there in the phrases that are selected by the new query features? I do not yet have answers for any of these questions.

## 12.   *Conclusions*

In this paper, I have introduced new features for keyphrase extraction. The new *query* features are inspired by the *keyphrase-frequency* feature of Frank *et al.* (1999). The query features use the PMI-IR algorithm to learn from a large corpus of unlabeled text. I have discussed the results of seven experiments:

1. Comparison of feature sets on the CSTR corpus: Query features improve performance. When trained and tested in the same domain, the query features perform significantly better than the baseline features, but not as well as the keyphrase features. Query features are not training-intensive. The query features used 130 training documents, compared to 1,300 for the

---

14. Although the AltaVista site has guidance on terms of use (http://www.altavista.com/sites/about/termsofuse), there is no discussion about how frequently a non-human user (a "robot") can issue queries. Robots have been crawling the Web for several years, and there are extensive guidelines for robot web crawlers (http://www.robotstxt.org/wc/exclusion.html), but I have not found any guidelines for robot web queriers. If the research described here is fruitful, then there will soon be a requirement for such guidelines.

   keyphrase features.
2. Generalization from the CSTR corpus to the LANL corpus: Query features are not domain-specific. When trained in one domain and then tested in another, the query features continue to perform better than the baseline features, but the performance of the keyphrase features drops below the baseline.
3. Evaluation of feature subsets: All of the twelve query features appear to be useful.
4. Relations among feature sets: Although the query features are conceptually similar to the keyphrase features, they result in substantially different output phrases.
5. Combining query features with the keyphrase feature: When the appropriate data are available, it is beneficial to combine the query features with the keyphrase feature. However, this hybrid shares the limitations of the keyphrase feature, domain-specificity and training-intensiveness.
6. Evaluating keyphrases by familiarity: The phrases extracted by the keyphrase feature set are highly "familiar" when the training and testing domains are the same, and more familiar than the baseline phrases when the training and testing domains are distinct. The phrases extracted by the query feature set are about as familiar as the baseline phrases when the training and testing domains are the same, and more familiar than the baseline and keyphrases phrases when the training and testing domains are distinct.
7. Evaluating keyphrases by searching: The phrases extracted by the keyphrase feature set appear to have a bias towards generality.

In summary, the main lesson is that the new features improve keyphrase extraction, yet they are neither domain-specific nor training-intensive.

## *Acknowledgments*

Thanks to the Kea group, Eibe Frank, Gordon Paynter, Ian Witten, Carl Gutwin, and Craig Nevill-Manning, for giving me a copy of the CSTR corpus, for releasing their Kea software under the GNU General Public License, and for sharing their results with me. Thanks to my colleague Alain Désilets for suggesting, by example, the idea of using a Web search engine as a source of input for an algorithm. Thanks to the developers and maintainers of the following search engines for permitting my software to send large numbers of queries to their search engines: AltaVista (for Web searching), New Zealand Digital Library (for searching the CSTR corpus), and the CERN Document Server (for searching the LANL corpus).

## *References*

Banko, M., Mittal, V., Kantrowitz, M., and Goldstein, J. (1999). Generating extraction-based summaries from hand-written summaries by aligning text spans. In *Proceedings of the Pacific Rim Conference on Computational Linguistics (PACLING-99).*

Church, K.W., Hanks, P. (1989). Word association norms, mutual information and lexicography. *Proceedings of the 27th Annual Conference of the Association of Computational Linguistics,* pp. 76-83.

Church, K.W., Gale, W., Hanks, P., Hindle, D. (1991). Using statistics in lexical analysis. In Uri Zernik (ed.), *Lexical Acquisition: Exploiting On-Line Resources to Build a Lexicon,* pp. 115-164. New Jersey: Lawrence Erlbaum.

Domingos, P., and Pazzani, M. (1997). On the optimality of the simple Bayesian classifier under zero-one loss. *Machine Learning,* 29, 103-130.

Dumais, S., Platt, J., Heckerman, D. and Sahami, M. (1998). Inductive learning algorithms and representations for text categorization. *Proceedings of the Seventh International Conference on Information and Knowledge Management,* pp. 148-155. ACM Press.

Edmundson, H.P. (1969). New methods in automatic extracting. *Journal of the Association for Computing Machinery,* 16 (2), 264-285.

Fayyad, U.M., and Irani, K.B. (1993). Multi-interval discretization of continuous-valued attributes for classification learning. In *Proceedings of 13th International Joint Conference on Artificial Intelligence (IJCAI-93),* pp. 1022-1027.

Feelders, A., and Verkooijen, W. (1995). Which method learns the most from data? Methodological issues in the analysis of comparative studies. *Fifth International Workshop on Artificial Intelligence and Statistics,* Ft. Lauderdale, Florida, pp. 219-225.

Field, B.J. (1975). Towards automatic indexing: Automatic assignment of controlled-language indexing and classification from free indexing. *Journal of Documentation,* 31 (4), 246-265.

Firth, J.R. (1957). A synopsis of linguistic theory 1930-1955. In *Studies in Linguistic Analysis,* pp. 1-32. Oxford: Philological Society. Reprinted in F.R. Palmer (ed.), *Selected Papers of J.R. Firth 1952-1959*, London: Longman (1968).

Frank, E., Paynter, G.W., Witten, I.H., Gutwin, C., and Nevill-Manning, C.G. (1999). Domain-specific keyphrase extraction. *Proceedings of the Sixteenth International Joint Conference on Artificial Intelligence (IJCAI-99)*, pp. 668-673. California: Morgan Kaufmann.

Furnas, G., Landauer, T., Gomez, L., & Dumais, S. (1987). The vocabulary problem in human-system communication. *Communications of the ACM,* 30, 964-971.

Gutwin, C., Paynter, G.W., Witten, I.H., Nevill-Manning, C.G., and Frank, E. (1999). Improving browsing in digital libraries with keyphrase indexes. *Journal of Decision Support Systems*, 27, 81-104.

Jones, S., and Paynter, G.W. (1999) Topic-based browsing within a digital library using keyphrases. *Proceedings of Digital Libraries 99 (DL'99)*, pp. 114-121. ACM Press.

Kupiec, J., Pedersen, J., and Chen, F. (1995). A trainable document summarizer. In E.A. Fox, P. Ingwersen, and R. Fidel, editors, *SIGIR-95: Proceedings of the 18th Annual International ACM SIGIR Conference on Research and Development in Information Retrieval*, pp. 68-73, New York: ACM.

Landauer, T.K., and Dumais, S.T. (1997). A solution to Plato's problem: The Latent Semantic Analysis theory of the acquisition, induction, and representation of knowledge. *Psychological Review*, 104: 211-240.

Leung, C.-H., and Kan, W.-K. (1997). A statistical learning approach to automatic indexing of controlled index terms. *Journal of the American Society for Information Science*, 48, 55-66.

Lovins, J.B. (1968). Development of a stemming algorithm. *Mechanical Translation and Computational Linguistics,* 11, 22-31.

Luhn, H.P. (1958). The automatic creation of literature abstracts. *I.B.M. Journal of Research and Development,* 2 (2), 159-165.

Manning, C.D., and Schütze, H. (1999). *Foundations of Statistical Natural Language Processing.* Cambridge, Massachusetts: MIT Press.

Martin, J., and Holte, R.C. (1998). Searching for content-based addresses on the World-Wide Web. *Proceedings of The Third ACM Conference on Digital Libraries (DL'98).*

Soderland, S., and Lehnert, W. (1994). Wrap-Up: A trainable discourse module for information extraction. *Journal of Artificial Intelligence Research,* 2, 131-158.

Sparck Jones, K. (1973). Does indexing exhaustivity matter? *Journal of the American Society for Information Science,* September-October, 313-316.

Turney, P.D., and Halasz, M. (1993), Contextual normalization applied to aircraft gas turbine engine diagnosis, *Journal of Applied Intelligence,* 3, 109-129.

Turney, P.D. (1997). *Extraction of Keyphrases from Text: Evaluation of Four Algorithms. National Research Council, Institute for Information Technology*, Technical Report ERB-1051.

Turney, P.D. (1999). *Learning to Extract Keyphrases from Text. National Research Council, Institute for Information Technology*, Technical Report ERB-1057.

Turney, P.D. (2000). Learning algorithms for keyphrase extraction. *Information Retrieval*, 2, 303-336.

Turney, P.D. (2001). Mining the Web for synonyms: PMI-IR versus LSA on TOEFL. *Proceedings of the Twelfth European Conference on Machine Learning (ECML-2001)*, Freiburg, Germany, pp. 491-502.

Turney, P.D. (2002). Answering subcognitive Turing Test questions: A reply to French. *Journal of Experimental and Theoretical Artificial Intelligence,* 13, 409-419.

van Rijsbergen, C.J. (1979). *Information Retrieval.* 2nd edition. London: Butterworths.

Whitley, D. (1989). The GENITOR algorithm and selective pressure. *Proceedings of the Third International Conference on Genetic Algorithms (ICGA-89)*, pp. 116-121. California: Morgan Kaufmann.

Witten, I.H., Paynter, G.W., Frank, E., Gutwin, C. and Nevill-Manning, C.G. (1999) KEA: Practical automatic keyphrase extraction. *Proceedings of Digital Libraries 99 (DL'99),* pp. 254-256. ACM Press.

Witten, I.H., Paynter, G.W., Frank, E., Gutwin, C., and Nevill-Manning, C.G. (2000). *KEA: Practical Automatic Keyphrase Extraction.* Working Paper 00/5, Department of Computer Science, The University of Waikato.